\documentclass[10pt,twocolumn,letterpaper]{article}
\usepackage{wrapfig}
\usepackage{booktabs}
\usepackage{multirow}
\usepackage{graphicx}
\usepackage{amssymb}
\usepackage{pifont}
\usepackage{{amsmath}}
\usepackage{graphicx}     % for \includegraphics
\usepackage{floatrow}     % for floatrow, \ffigbox, \CenterFloatBoxes
\usepackage[font=small]{caption}
\usepackage{tabularx}
\usepackage{makecell}
\usepackage{placeins}

\usepackage{floatrow}
\newfloatcommand{capbtabbox}{table}[][\FBwidth] 
\usepackage{caption}
\usepackage{subcaption}

\usepackage[pagenumbers]{cvpr} % To force page numbers, e.g. for an arXiv version

\definecolor{cvprblue}{rgb}{0.21,0.49,0.74}
\usepackage[pagebackref,breaklinks,colorlinks,allcolors=cvprblue]{hyperref}

\def\confName{CVPR}
\def\confYear{2026}

\title{Live Interactive Training for Video Segmentation}

\author{
Xinyu Yang \quad
Haozheng Yu \quad
Yihong Sun \quad
Bharath Hariharan \quad
Jennifer J. Sun\\ \\
Cornell University
}

\begin{document}
\maketitle
\begin{abstract}
Interactive video segmentation often requires many user interventions for robust performance in challenging scenarios (e.g., occlusions, object separations, camouflage, etc.).
Yet, even state-of-the-art models like SAM2 use corrections only for immediate fixes without \textit{learning} from this feedback, leading to inefficient, repetitive user effort. 
To address this, we introduce Live Interactive Training (LIT), a novel framework for prompt-based visual systems where models also learn online from human corrections at inference time. 
Our primary instantiation, LIT-LoRA, implements this by continually updating a lightweight LoRA module on-the-fly. When a user provides a correction, this module is rapidly trained on that feedback, allowing the vision system to improve performance on subsequent frames of the same video.
Leveraging the core principles of LIT, our LIT-LoRA implementation achieves an average 18-34\%  reduction in total corrections on challenging video segmentation benchmarks, with a negligible training overhead of $\sim$0.5s per correction. We further demonstrate its generality by successfully adapting it to other segmentation models and extending it to CLIP-based fine-grained image classification. Our work highlights the promise of live adaptation to transform interactive tools and significantly reduce redundant human effort in complex visual tasks. Project: \href{https://youngxinyu1802.github.io/projects/LIT/}{https://youngxinyu1802.github.io/projects/LIT/}.

\end{abstract}

\section{Introduction}
\label{sec:intro}

\begin{figure}[h]
    \centering
    \includegraphics[width=1.0\linewidth]{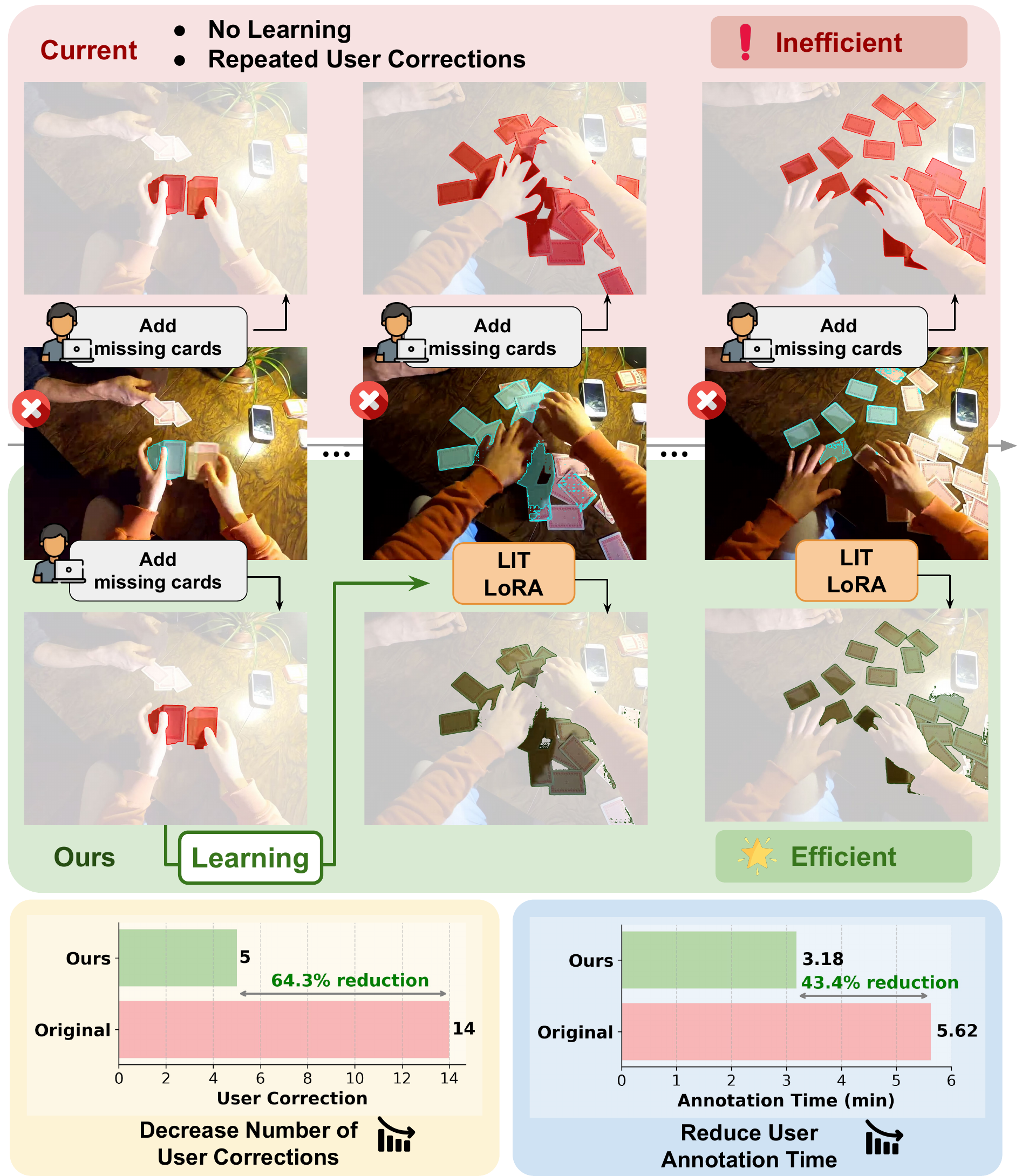}
    \vspace{-4mm}
    \caption{\small \textbf{Comparison between the current non-learning system and our LIT-LoRA approach.} The current system (top) does not learn from user feedback, 
    leading to the same errors to reappear and requiring repeated corrections (e.g., 14 prompts to add the missing cards), which leads to substantial annotation time (e.g., 5.62 mins). In contrast, our LIT-LoRA method continuously adapts to user correction input and generalizes to similar future errors, reducing the number of required corrections (e.g., down to 5) and user annotation time (e.g., down to 3.18 mins). 
    }
    \label{fig:fig1}
\end{figure}

Modern vision foundation models~\cite{sam2,sam} have introduced powerful new capabilities for interacting with users, such as generating video segmentation masks from simple clicks or boxes. 
However, despite this transformative potential, achieving robust performance in complex real-world scenarios remains challenging.
Persistent issues such as occlusions~\cite{dam4sam, sam2long}, look-alike objects~\cite{camsam2}, non-rigid deformations~\cite{vscos, vost}, and small instances~\cite{rmem, samurai, sam2long} frequently lead to errors that demand substantial human intervention within a single video.

This human-model collaboration, however, faces a critical limitation: even state-of-the-art models like SAM2 primarily leverage user corrections for immediate prediction refinement. Such feedback is often applied temporarily or cached in memory without leading to fundamental model adaptation. Consequently, the system does not truly \textit{learn} or generalize from these valuable interactions to subsequent frames, forcing users into an inefficient and frustrating cycle of repeatedly correcting the same types of errors within the same video. This substantially limits overall workflow efficiency and long-term adaptability. 

Figure~\ref{fig:fig1} illustrates this: segmenting challenging subjects like the separating cards with SAM2 alone requires numerous manual interventions (e.g., 14 corrections taking more than 5 minutes) due to such recurring failures. Ideally, a system would learn from these initial corrections to autonomously handle subsequent, similar challenges, thereby significantly reducing the corrective burden. Our approach demonstrates this capability, requiring only 5 corrections in the same scenario and reducing the annotation time.

To address the challenge of recurring corrections, there is a need for a framework for interactive vision models to learn and adapt live from user guidance. 
The rise of powerful vision foundation models and the development of parameter-efficient adaptation techniques make such online learning~\cite{online, onlinemeta} increasingly practical. 
We introduce Live Interactive Training (LIT) to realize this vision. LIT operationalizes principles of online learning specifically for the demands of prompt-based interactive visual systems  (e.g., those guided by user clicks, boxes, or masks), enabling continuous adaptation from human corrections directly during inference. We initiate LIT's potential within video segmentation, a domain where user corrections are expensive and which demands accurate, adaptable, and efficient outcomes. 

To realize the LIT framework, we introduce LIT-LoRA, a lightweight and modular implementation. LIT-LoRA trains compact LoRA modules online, leveraging user corrections in real-time.  
This approach allows the underlying foundation model to capture, generalize, and dynamically adapt to user-provided feedback.
We extensively evaluate LIT-LoRA using SAM2 as a leading video segmentation exemplar. To rigorously benchmark the efficiency, we evaluate our method using a controlled, reproducible protocol with synthetic user corrections, a standard practice in interactive segmentation~\cite{sam,sam2,mivos,eva-vos}.
To show generality, we further extend LIT to a CLIP-based image classification task, where user feedback is textual, and observe consistent improvements. We lay the foundation for interactive visual systems that are more efficient and adaptive, towards human-AI collaboration in complex real-world settings.

To summarize, our contributions are:
\begin{itemize}
    \item We present LIT, a novel framework for prompt-based visual systems that enables models to learn and adapt from human corrections during inference.
    \item We propose LIT-LoRA, a practical and lightweight implementation of LIT that demonstrates significant reductions in user effort (18-34\% fewer corrections in challenging video segmentation) with negligible ($\sim$0.5s) overhead.
    \item We demonstrate the generality of the LIT framework, showing it is model-agnostic (improving multiple SAM2 variants) and task-agnostic (extending to CLIP-based image classification).
\end{itemize}
\section{Related Work}

\paragraph{Video Object Segmentation.}
The task of video object segmentation (VOS) is to segment an object throughout a video given its first-frame mask. One of the main approaches uses memory-based representations to propagate features across frames~\cite{xmem, rmem, cutie}. Building on this paradigm, SAM2~\cite{sam2} extends SAM~\cite{sam} to videos by incorporating a memory bank of object features. While achieving strong zero-shot generalization across benchmarks, its performance drops in challenging scenarios~\cite{hong2024lvosbenchmarklargescalelongterm, vost,vscos, mocamask, CAD, sun2025tracking}.
To address the limitations, several recent studies have extended SAM2 with additional mechanisms.
Many methods focus on enhancing memory design. For example, SAM2Long~\cite{sam2long} introduces a tree-based memory retrieval strategy to reduce error propagation; SAMURAI~\cite{samurai} incorporates motion cues for more stable multi-object tracking; and DAM4SAM~\cite{dam4sam} adopts a distractor-aware memory update to improve robustness among similar objects.
Other approaches modify the feature representation of SAM2 by introducing new learnable tokens (e.g. CAMSAM2~\cite{camsam2} and HQ-SAM2~\cite{sam_hq}).
There are also efforts to adapt SAM2 to specific domains through finetuning~\cite{sam2adapter,surgicalsam2,polysam2}.
Despite these advances, they primarily focus on model or domain adaptation, and overlook a powerful, inherent capability of SAM2: user correction. 
Our work investigates a complementary approach: instead of modifying the core model, we leverage its interactive capability to enhance performance, specifically investigating how to integrate user feedback effectively to minimize repetitive user effort.

\paragraph{User-interactive Visual Systems.}
Human-in-the-loop collaboration enables models to achieve greater performance and usability through interactive support. Systems such as PromptCharm~\cite{wang2024promptcharm}, DesignPrompt~\cite{peng2024designprompt}, and MagicQuill~\cite{liu2025magicquill} explore how generative models can better interpret user sketches and strokes for interactive image editing. In image and video object segmentation, visual prompting similarly leverages user-provided cues (e.g., points, boxes, masks) to guide segmentation~\cite{sam, sam2, mivos, eva-vos, lazy-xeme}. However, existing works focus on improving how models interpret user intent efficiently, rather than enabling models to learn from user interactions to correct future errors.

\paragraph{Parameter-Efficient Fine-Tuning.}
With the rise of large foundation models~\cite{dinov3, clip, dubey2024llama, videomae, bai2023qwen, zhao2024videoprism}, full fine-tuning has become prohibitively expensive in memory, storage, and latency. Parameter-Efficient Fine-Tuning (PEFT) mitigates this cost by freezing the backbone and updating only a small set of parameters or lightweight modules, enabling efficient adaptation to downstream tasks.
This lightweight structure and minimal trainable footprint make PEFT models well-suited for our online interactive tasks where low latency and small memory overhead are key requirements. Among the various PEFT methods~\cite{peft, rosa, peft-fourier, ni2024pace, hu2023vl, hu2023llm}, we adopt LoRA~\cite{hu2022lora} in our framework for its efficiency and ease of use, while our framework is inherently compatible with other PEFT techniques. 
LoRA injects trainable low-rank matrices into the Transformer layers while keeping the original weights frozen, achieving strong adaptation performance across diverse domains~\cite{cliplora, convlora, cellseg1, icedit}.  
However, most PEFT methods are trained offline on static, supervised datasets~\cite{wang2022adamix,sung2022vl,li2024adapter}. Even more recent work that explores online adaptation with LoRA focuses on continual learning from a stream of labeled data~\cite{wei2025online, he2025cl}.
Our framework employs LoRA for live, user-driven adaptation, enabling real-time model updates from interactive user feedback while keeping the training lightweight.

\paragraph{Model Adaptation Paradigm.}

Several model adaptation paradigms have emerged in recent years, such as Test-Time Training (TTT)~\cite{TTT, TTTLLM, TTTMAE}, Continual Test-Time Adaptation (CTTA)~\cite{CTTA, controlCTTA}, and Online Learning~\cite{shalev2012online, qian2024efficient}.
TTT adapts a pre-trained model using unlabeled test samples at inference time via self-supervised objectives, enabling limited domain adaptation but assuming a static test set. OSVOS~\cite{osvos} follows this paradigm by fine-tuning on the first video frame to adapt the VOS model.
CTTA extends this idea to streaming inputs, continuously updating models to shifting distributions using pseudo-labels, as demonstrated in VOS by OnAVOS~\cite{onavos}.
Online Learning~\cite{wei2023online} instead incrementally updates models using ground-truth labels as they arrive in the data stream.
Our framework, LIT-LoRA, can be viewed as a \emph{user-feedback-driven variant of online learning} that operates at inference time. While it shares CTTA's streaming and deployment-time nature, it differs critically in supervision: adaptation is guided by human corrections rather than pseudo-labels. This interactive signal turns the process into a lightweight, human-in-the-loop continual adaptation scheme, combining the immediacy of online learning with the practicality of inference-time adaptation. Furthermore, unlike continuous adaptation methods for single-image segmentation~\cite{atanyan2024continuous, continuous}, LIT-LoRA adapts at the level of the data stream (e.g., a video containing multiple frames), enabling the correction to benefit later samples that share similar challenges.
\section{Method}
\label{sec:method}

\begin{figure*}[!t]
    \centering
\includegraphics[width=1\linewidth]{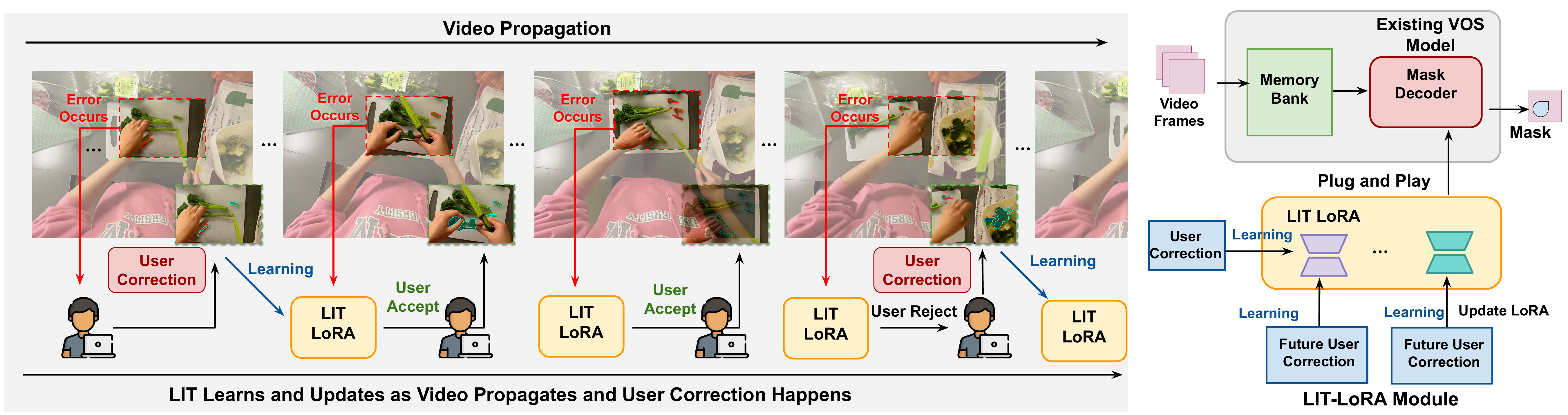}
\vspace{-6mm}
\caption{\small \textit{Left:} \textbf{Overview of the LIT-LoRA framework on VOS}. As the video progresses, segmentation errors may arise. When the user provides a correction (which can be time-consuming), the correction is used to train a LoRA module on-the-fly. The LoRA module is then consulted for later errors: if its prediction meets the validation criterion, it is accepted to correct the error; otherwise, the adapter is further refined using the latest correction. \textit{Right:} \textbf{LIT LoRA module illustration}.
}
    \label{fig:lora_overview}
\end{figure*}
\label{sec:related}

We propose Live Interactive Training (LIT), a framework designed to enable lightweight, real-time adaptation for user corrections in interactive visual systems. While the framework is designed to be task- and model-agnostic, we initialize it on SAM2 (Section~\ref{sec:sam2}), as it serves as a strong, representative system for user-interactive segmentation. In the following, we first present the overall LIT framework (Section~\ref{sec:LIT}) and then detail its specific implementation, LIT-LoRA (Section~\ref{sec:LIT-LoRA}).

\subsection{SAM2 Preliminaries}
\label{sec:sam2}

SAM2~\cite{sam2} is an exemplar model for promptable visual segmentation (PVS) in temporally dynamic settings. In the PVS framework, users can provide prompts, such as positive/negative clicks, bounding boxes, or masks, on the video frame to initialize object segmentation. Once a prompt is received, the model immediately returns a segmentation mask for that frame and then propagates the object representation throughout the video to generate a masklet (i.e. the predicted object mask on every frame). 

To maintain interactivity, additional prompts can be provided at any time to refine the video-wide prediction further. Architecturally, SAM2 achieves this interaction by augmenting the standard SAM architecture with a memory bank that stores object features from previous propagation. While this promptable design offers flexibility, the model does not truly learn from these user interactions. Corrections are merely stored as conditional inputs or as recent-frame context in the memory bank for the model. However, since the model's parameters remain frozen, it cannot internalize or generalize from these corrections. As a result, similar errors can persist in future frames, limiting the system's ability to improve over time.

\subsection{LIT Framework}
\label{sec:LIT}

We introduce the \textbf{LIT (Live Interactive Training)} framework for building interactive visual systems that adapt continuously from user feedback during inference. Unlike conventional approaches that rely on static, pre-trained models or batch fine-tuning, LIT is a \emph{user-feedback-driven variant of online learning} that operates at inference time. It emphasizes real-time and user-driven adaptation, enabling models to improve performance dynamically in response to user feedback.

In the LIT setting, data arrive as a stream
$
\{x_t\}_{t=1}^{T},
$
where each data $x_t$ (e.g., an image or a video frame) is processed sequentially within the stream. For stable online adaptation, the stream is conceptually divided into coherent groups, such as individual videos or subsets of samples that share similar visual or semantic characteristics.
As each sample is processed, the model produces an initial prediction
$
\hat{y}_t = f_{\theta,\,\phi_t}(x_t),
$
where $\theta$ denotes frozen backbone parameters and $\phi_t$ denotes the lightweight, trainable adapter active at time $t$. 

When the user identifies an error and provides a correction $y_t^{\ast}$, that correction is immediately treated as a supervision signal. This signal is used to train a lightweight, parameter-efficient module $\phi_t$ in real-time. The adapter is updated via
$$
\phi_{t+1}
\leftarrow
\phi_t - \eta \,\nabla_{\phi_t}
\mathcal{L}\big(f_{\theta,\,\phi_t}(x_t),\, y_t^\ast\big),
$$
where $\mathcal{L}$ is a task-specific loss and $\eta$ is a small learning rate enabling stable real-time updates.

This adapted module is then used to improve predictions for subsequent data in the same stream:
$$
{y}^\phi_{t'} = f_{\theta,\,\phi_{t'}}(x_{t'}), \quad t' > t.
$$

The adapter is maintained and incrementally refined within each stream group, enabling it to accumulate group-specific knowledge and correct recurring failure patterns that stem from shared visual or semantic structures. When the system transitions to a new group, the adapter is reinitialized.

This update–predict cycle repeats whenever new corrections arrive, forming a closed-loop system in which the model continually incorporates user guidance.  As a result, LIT progressively reduces repeated errors and improves system efficiency during the same inference session, all without retraining the full backbone.

The key features of the LIT framework include:
\begin{itemize}
    \setlength\itemindent{0em}         
    \item \textbf{Live training:} The model supports low-latency updates during inference.
    \item \textbf{Interactive learning:} The system is able to respond immediately to user feedback and learn from this interactive result.
    \item \textbf{Online improvement:} The model continuously incorporates feedback to refine predictions over time, improving its ability as the system proceeds.
\end{itemize}

\subsection{LIT-LoRA on VOS}
\label{sec:LIT-LoRA}
While the LIT framework is designed to be task- and model-agnostic, we instantiate it on the video object segmentation (VOS) task using LoRA (denoted as $\mathcal{A}_t$) as the lightweight learning module ($\phi_t$ in the LIT framework), forming \textbf{LIT-LoRA}. In this setup, the LoRA adapters are attached to a frozen segmentation backbone ($\theta$) and updated online in response to user corrections. This design enables real-time learning with minimal overhead and allows the model to transfer correction patterns to future errors, thereby reducing redundant user effort and improving segmentation quality over time.

As illustrated in Figure~\ref{fig:lora_overview}, LIT-LoRA operates as a live, user-driven adaptation loop. During inference, the model sequentially processes incoming video frames and may produce segmentation errors. When an error is identified, the user provides a correction for that frame, which serves as an immediate supervision signal. The LoRA adapter is then trained on-the-fly to integrate this correction. For future error cases, the updated adapter generates refined predictions; if the predicted mask meets the quality criterion, it is accepted, otherwise the user provides another correction and the adapter is updated again. This continual loop allows the model to adapt dynamically within the same inference session, effectively closing the gap between inference and training. 
We describe the detailed process below. 

\paragraph{Interactive Error Detection and Correction}  
In the LIT framework, the adaptation loop is initiated by an error trigger. In a real-world interactive scenario, this trigger is a manual user correction (e.g., clicks or masks) provided when the user identifies a segmentation failure. This correction, $M_t^{\text{corr}}$, serves as feedback for model adaptation. 

\paragraph{Live Model Updating from User Corrections}  
A lightweight LoRA module $\mathcal{A}_t$ is trained whenever a correction is received. It takes the visual embedding of the current frame $F_t$ together with the corresponding memory bank information to obtain fused features $x_t$, and outputs a refined mask prediction
${M}^{\mathcal{A}}_t = f_{\theta,\mathcal{A}_t}(x_t)$.
The adapter is optimized using a standard segmentation loss
$$\mathcal{L} = \mathcal{L}_{\text{seg}}({M}_t^{\mathcal{A}}, M_t^{\text{corr}}).$$
We follow the SAM2 training setup, which combines focal and Dice losses with a weighting ratio of $20{:}1$ to balance pixel-level accuracy and region-level consistency.
Since only a small number of low-rank parameters in $\mathcal{A}_{t}$ are updated, the optimization converges typically under one second and incurs minimal computational or memory overhead. The updated adapter is immediately applied to subsequent errors, enabling the system to incorporate user feedback in real time and progressively refine segmentation quality across the video.

\paragraph{Propagation and Validation of Updates}  

For a future frame $F_{t'}$ where an error occurs, the system employs the updated LoRA adapter $\mathcal{A}_{t'}$ to produce a refined mask prediction $M_{t'}^{\mathcal{A}}= f_{\theta,\mathcal{A}_{t'}}(x_{t'})$. This refined mask is presented to the user to validate: 
\begin{itemize}
    \item If the user accepts this prediction (i.e., by not providing another correction), the prediction is accepted as the final output for $F_{t'}$. It replaces the previous result and can be stored in the memory bank to enhance future propagation.
    \item If the user identifies a new failure and provides a new correction $M_{t'}^{\text{corr}}$, this is treated as a new error event.
\end{itemize}

In the second case, the system flags an error, and the LoRA module $\mathcal{A}_{t'}$ is then incrementally updated using this new correction signal, enabling the model to refine its parameters in real time. This continual loop of user validation and model adaptation allows the system to dynamically adjust to new frames, maintain robust segmentation performance, and minimize the number of user interventions required for high-quality results.
\section{Experiments}
\label{sec:experiments}
\begin{table*}[t]
\centering
\setlength{\tabcolsep}{4pt}
\renewcommand{\arraystretch}{0.95}
\footnotesize
\caption{\textbf{Comparison of user corrections and annotation time across datasets under different IoU thresholds.}
LIT consistently reduces both user corrections and annotation time at $\tau_{\mathrm{IoU}}$ = 0.5 and 0.75.}
\vspace{-1mm}
\resizebox{\textwidth}{!}{
\begin{tabular}{llccccc!{\vrule width 0.8pt}ccccc}
\toprule
 &  & \multicolumn{5}{c!{\vrule width 0.8pt}}{\textbf{(a) Average user corrections per video}} 
 & \multicolumn{5}{c}{\textbf{(b) Average annotation time per video (min)}} \\
\cmidrule(lr){3-7}\cmidrule(lr){8-12}

\textbf{$\tau_{\mathrm{IoU}}$} & \textbf{Method}
& \textbf{VOST} & \textbf{LVOSv2} & \textbf{MOSEv2} & \textbf{SA-V Val} & \textbf{SA-V Test}
& \textbf{VOST} & \textbf{LVOSv2} & \textbf{MOSEv2} & \textbf{SA-V Val} & \textbf{SA-V Test} \\
\midrule

\multirow{3}{*}{0.5}
& Original
& 27.43 & 33.59 & 31.78 & 20.66 & 20.90
& 18.42 & 14.83 & 22.49 & 12.90 & 13.09 \\

& LIT
& 18.24 & 25.75 & 25.99 & 16.91 & 16.23
& 12.91 & 11.86 & 18.00 & 10.01 & 10.73 \\

& Reduced
& \textcolor{green!60!black}{$\downarrow$\textbf{33.51\%}}
& \textcolor{green!60!black}{$\downarrow$\textbf{23.35\%}}
& \textcolor{green!60!black}{$\downarrow$\textbf{18.22\%}}
& \textcolor{green!60!black}{$\downarrow$\textbf{18.16\%}}
& \textcolor{green!60!black}{$\downarrow$\textbf{22.35\%}}
& \textcolor{green!60!black}{$\downarrow$\textbf{29.94\%}}
& \textcolor{green!60!black}{$\downarrow$\textbf{20.03\%}}
& \textcolor{green!60!black}{$\downarrow$\textbf{19.98\%}}
& \textcolor{green!60!black}{$\downarrow$\textbf{22.44\%}}
& \textcolor{green!60!black}{$\downarrow$\textbf{18.01\%}} \\

\midrule

\multirow{3}{*}{0.75}
& Original
& 40.40 & 56.32 & 41.47 & 27.48 & 29.83
& 39.47 & 46.15 & 41.96 & 26.18 & 28.16 \\

& LIT
& 30.85 & 45.00 & 34.43 & 22.55 & 24.50
& 31.01 & 36.65 & 33.15 & 21.13 & 23.00 \\

& Reduced
& \textcolor{green!60!black}{$\downarrow$\textbf{23.65\%}}
& \textcolor{green!60!black}{$\downarrow$\textbf{20.10\%}}
& \textcolor{green!60!black}{$\downarrow$\textbf{16.98\%}}
& \textcolor{green!60!black}{$\downarrow$\textbf{17.94\%}}
& \textcolor{green!60!black}{$\downarrow$\textbf{17.89\%}}
& \textcolor{green!60!black}{$\downarrow$\textbf{21.43\%}}
& \textcolor{green!60!black}{$\downarrow$\textbf{20.58\%}}
& \textcolor{green!60!black}{$\downarrow$\textbf{21.00\%}}
& \textcolor{green!60!black}{$\downarrow$\textbf{19.29\%}}
& \textcolor{green!60!black}{$\downarrow$\textbf{18.32\%}} \\

\bottomrule
\end{tabular}
}
\vspace{-2mm}
\label{tab:iou_comparison}
\end{table*}

\begin{figure*}[ht]
  \begin{center}
    \includegraphics[width=1.0\textwidth]{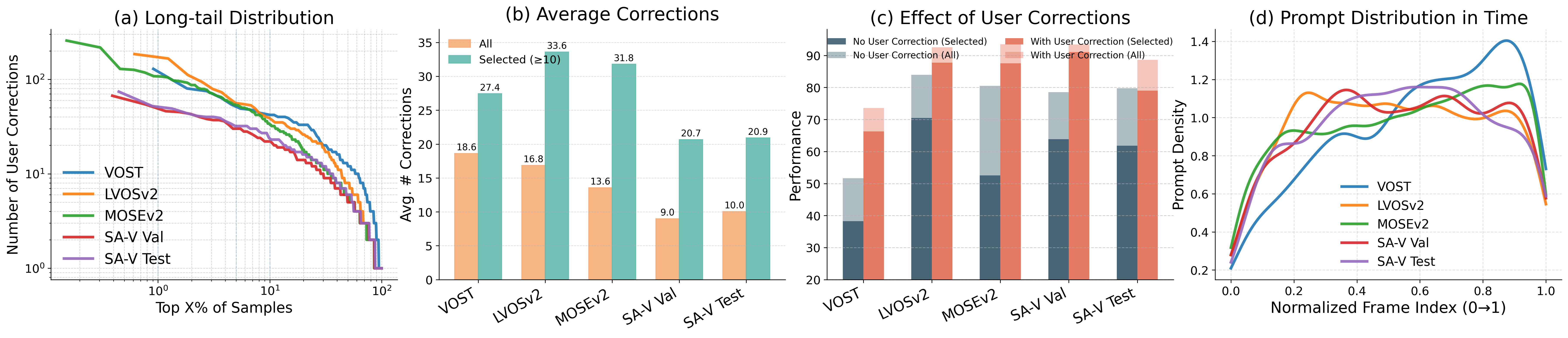}
  \end{center}
  \vspace{-6mm}
  \caption{\textbf{User interaction patterns and the impact across datasets.}
(a) The number of user corrections follows a clear long-tailed distribution: a small fraction of challenging videos accounts for the majority of interactions.
(b) The challenging cases ($\geq$ 10 corrections) require substantially more user inputs than the dataset average.
(c) User feedback consistently improves segmentation performance, especially for the challenging subset.
(d) Corrections are not uniformly distributed in time; most prompts occur in the early to late portions of each sequence, indicating the recurrence of errors. 
}
  \label{fig:empirical_analysis}
\end{figure*}

\begin{figure*}[ht]
    \centering
    \includegraphics[width=1.0\linewidth]{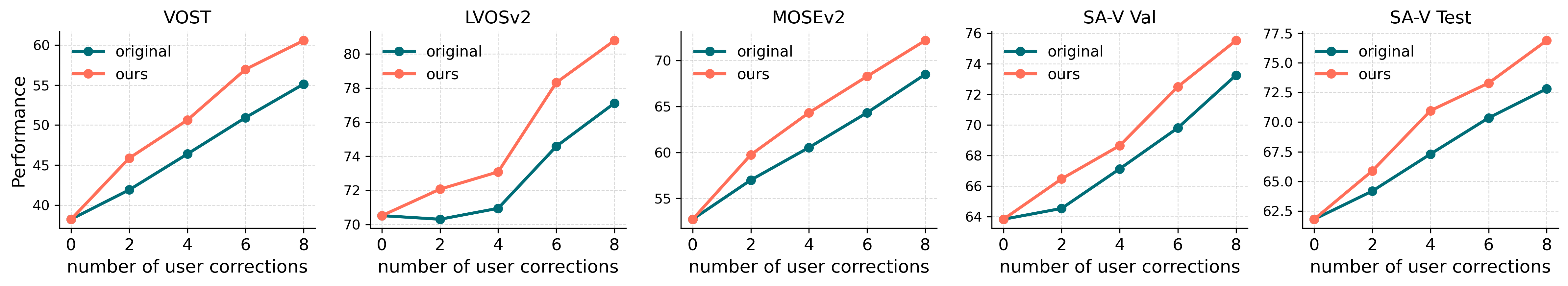}  
\caption{\small \textbf{Performance under different numbers of user corrections.}} 

    \label{fig:performance}
\end{figure*}
\begin{figure*}[ht]
    \centering
    \includegraphics[width=1.0\linewidth]{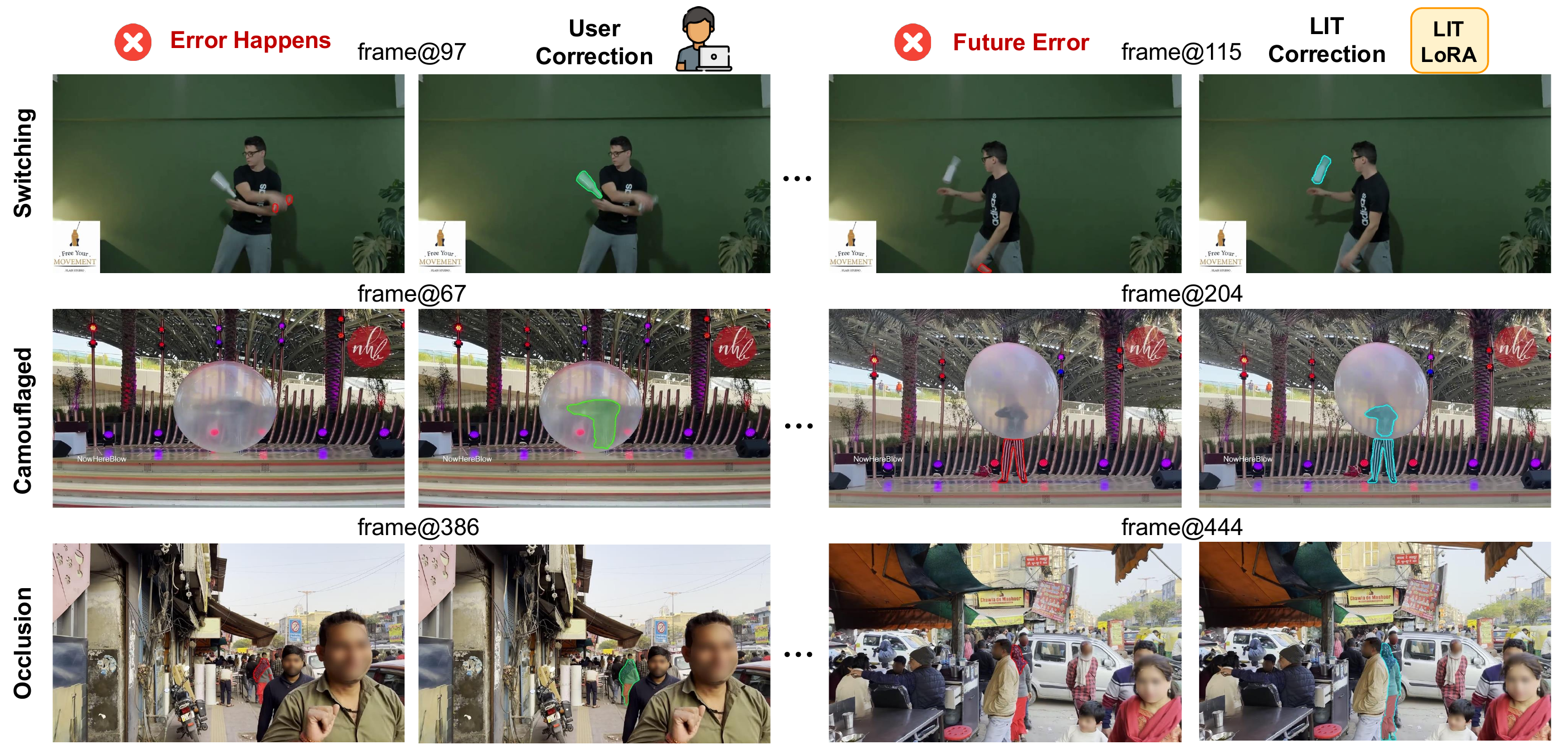}    
    \vspace{-6mm}
    \caption{\small \textbf{Qualitative results.}
    }
    \label{fig:qualitative}
\end{figure*}
\subsection{Experiment Setup}
\label{sec:exp_setup}

We evaluate LIT-LoRA under an interactive online evaluation protocol following the default settings of SAM2~\cite{sam2}. The model processes video frames sequentially in a single forward pass, where each frame is visited once and never revisited, reflecting the real-time nature of interactive usage. For each frame $F_t$, the model predicts a segmentation mask $M_t$. The prediction quality is assessed by $\text{IoU}(M_t, M_t^{\text{gt}})$. When this score falls below a predefined threshold $\tau_{\mathrm{IoU}}$, an error event is triggered and the system enters a correction phase. Unless otherwise specified, we use $\tau_{\mathrm{IoU}}=0.5$ as the default error-trigger threshold, corresponding to visibly noticeable failures and aligning with the standard AP@50 evaluation threshold.

Following standard practice in interactive segmentation~\cite{sam, sam2, mivos, eva-vos}, we employ synthetic user corrections for control and reproducibility.
Corrections are simulated as follows: (1) the user first provides up to three clicks at error centers to locally refine the mask; (2) if the IoU remains below $\tau_{\mathrm{IoU}}$, a full ground-truth mask is supplied instead. This hybrid strategy balances efficiency and effectiveness, since clicks enable rapid local adjustments, while full masks recover complex errors that cannot be easily fixed by clicks.

During correction, the simulated user mask $M_t^{\text{corr}}$ supervises online updates of the LoRA adapter $\mathcal{A}$ while the backbone remains frozen. 
For subsequent frames $F_{t'}$, the updated adapter produces refined predictions $M_{t'}^{\mathcal{A}}$. 
Masks with $\mathrm{IoU} \ge \tau_{\mathrm{IoU}}$ are accepted and stored in the memory bank for future propagation; otherwise, further corrections are applied.

In our LoRA training, the LoRA is configured with rank of 4, $\alpha=4$, dropout of 0.1, and a learning rate of $1\times10^{-4}$. The LoRA is injected to each layer of the mask decoder, and each LoRA is trained for 40 epochs with early stop.

\paragraph{Datasets and Evaluation}
We evaluate our method on four challenging VOS benchmarks: \textit{VOST}~\cite{vost}, \textit{LVOSv2}~\cite{hong2024lvosbenchmarklargescalelongterm}, \textit{MOSEv2}~\cite{ding2025mosev2} and \textit{SA-V}~\cite{sam2} test and val set. These benchmarks include challenging scenarios such as object separation, heavy occlusion, and long video sequences, which often require frequent user corrections. 
Detailed dataset descriptions are provided in Appendix~\ref{sec:supp_detail}.

We measure the number of corrections required to satisfy the target quality threshold. 
For segmentation performance, we follow standard VOS metrics: $\mathcal{J}\& \mathcal{F}$, which combines region similarity ($\mathcal{J}$, IoU) and contour accuracy ($\mathcal{F}$).
For VOST, we follow the official evaluation protocol and report $\mathcal{J}$ only. 
To estimate user interaction time, we adopt annotation time statistics from SAM2~\cite{sam2} and EVA-VOS~\cite{eva-vos}: localization takes $T_{\mathrm{loc}} = 1$ sec and each click requires $T_{\mathrm{click}} = 1.5$ sec, while a full-mask annotation takes $80$ sec.

\subsection{Empirical Analysis of User Interaction}

We begin by examining how user corrections occur in promptable video segmentation systems. Figure~\ref{fig:empirical_analysis} analyzes user correction patterns across different VOS benchmarks. We observe a long-tailed distribution of corrections (Figure~\ref{fig:empirical_analysis} (a)): while many videos require few interventions, a small subset of challenging sequences accounts for most of the total user effort. Correspondingly, the average number of corrections is modest, but the high-interaction subset ($\geq
10$ corrections) requires up to 2.3 times more effort (Figure~\ref{fig:empirical_analysis} (b)).  These are also the cases where user corrections yield the largest performance improvements (Figure~\ref{fig:empirical_analysis} (c)), meaning feedback is most valuable precisely where it is most costly. Since these high-interaction cases dominate both user burden and observable benefit, our LIT-LoRA experiments primarily target this challenging subset. Additionally, corrections are temporally concentrated at initialization and again later in the sequence due to temporal drift (Figure~\ref{fig:empirical_analysis} (d)), showing that errors reappear over time. This motivates converting user feedback into online adaptation to reduce repeated user interaction effort.

\subsection{Main Results of LIT-LoRA}

\paragraph{Reducing user corrections and annotation time}

Table~\ref{tab:iou_comparison} (a) (top)
shows the average number of user corrections required for each dataset, where each frame is required to have an IoU greater than $\tau_{\mathrm{IoU}}=0.5$.
Across all four datasets, our method reduces the user correction by 18\% to 34\%. The largest gains appear on VOST, which differs from SAM2's typical behavior because objects often split into multiple parts while SAM2 usually segments only one.

We further evaluate annotation cost in terms of total simulated annotation time as defined in Section~\ref{sec:exp_setup}, including the training and inference latency introduced by the LIT-LoRA module. As reported in Table~\ref{tab:iou_comparison} (b), LIT-LoRA reduces total annotation time by an average of 22.1\% across datasets. With a 2–5 minute reduction per video, it can save hours of annotation time for the entire dataset.

We additionally evaluate LIT-LoRA under a stricter quality requirement ($\tau_{\mathrm{IoU}} = 0.75$), where higher mask precision is demanded at every frame. As shown in Table~\ref{tab:iou_comparison} (bottom), LIT-LoRA consistently reduces user corrections and annotation time by 17–24\% across datasets. Although the relative reduction is smaller than under $\tau_{\mathrm{IoU}} = 0.5$, which is expected since stricter thresholds reject corrections that would otherwise be accepted, higher precision requirements markedly increase absolute user effort, whereas the computational overhead of LIT-LoRA remains constant at $\sim$0.5s per update. As a result, the absolute amount of user effort saved by LIT-LoRA remains significant under stricter requirements, highlighting a favorable trade-off between lightweight computation and human annotation cost.

\paragraph{Improving accuracy when fixing user corrections}
To assess how effectively each correction improves segmentation, we compare our method with the baseline by allowing up to a fixed number of user corrections per video and evaluating the resulting propagated performance. This setup simulates a realistic interactive scenario in which the user’s correction budget is limited.
As shown in Figure~\ref{fig:performance}, our method consistently achieves higher performance than the baseline under the same number of corrections. This validates the advantage of our approach: rather than applying corrections in a static manner, we adapt the model itself using lightweight online training, which allows each correction to improve not only the current frame but also generalize to future errors. 

\paragraph{Lightweight overhead}
We evaluate the runtime of LIT-LoRA on an RTX Ada 6000 GPU. The LoRA module introduces only 35K trainable parameters, ($\sim0.01\%$ of full-model fine-tuning), and each correction requires approximately $0.5 \pm 0.2$ s of online training. 
By comparison, the dominant cost in interactive segmentation remains the human input: annotating a single object typically takes about 1 sec to locate and 1.5 sec per click according to SAM2, and annotating a difficult object with a full mask can take up to 80 sec~\cite{eva-vos}. Under our hybrid correction strategy, the average per-correction annotation time is 32.1 sec at $\tau_{\mathrm{IoU}} = 0.5$ and 46.6 sec at $\tau_{\mathrm{IoU}} = 0.75$.
These results show that the training overhead of LIT-LoRA is negligible relative to human annotation. Consequently, the system maintains its lightweight, real-time responsiveness and is well-suited for interactive workflows.

\paragraph{Qualitative results} Figure~\ref{fig:qualitative} shows qualitative examples of our LIT-LoRA corrections when error occurs. Using prior correction signals, LIT-LoRA resolves recurring errors in later frames, including object switching, camouflaged objects, and occlusions.

\paragraph{User study} To validate our method with real human annotators, we conducted a small-scale user study. We built an annotation interface on top of the SAM2 demo with online correction and LIT-LoRA, based on point corrections.
We recruited 6 volunteers, who were first given time to familiarize themselves with the GUI. Each participant annotated 8 randomly selected videos from the VOST dataset (one used for familiarization) using both the baseline and our method in randomized order.
The study follows the same online protocol: users pause and correct immediately once they see an error.
To avoid familiarity bias, users reviewed the ground-truth masks before each session. 
We measure both correction number and correction time, which includes pausing, locating errors, and making corrections. We observe consistent reductions across users in correction number (\textbf{41.92\%}) and correction time (\textbf{23.04\%}). Note that the study used point corrections rather than masks, the relative time savings will be greater with more intensive corrections.

\begin{table}[ht] \centering \setlength{\tabcolsep}{4pt} \renewcommand{\arraystretch}{0.95} \footnotesize \caption{\textbf{User correction reduction across models and tasks.}} \vspace{-1mm} \resizebox{\textwidth}{!}{ 
    \begin{tabular}{@{}lcc!{\vrule width 1pt}ccc@{}}
    \toprule
    \multicolumn{3}{c!{\vrule width 1pt}}{\textbf{(a) VOS}} &
    \multicolumn{3}{c}{\textbf{(b) Fine-grained image classification}} \\
    \cmidrule(lr){1-3}\cmidrule(lr){4-6}
    \textbf{Dataset} & \multicolumn{2}{c!{\vrule width 1pt}}{\textbf{VOST}} &
    \textbf{CUB-200-2011} & \textbf{Stanford Cars} & \textbf{SUN397} \\
    \textbf{Model} & \textbf{DAM4SAM} & \textbf{SAMURAI} &
    \multicolumn{3}{c}{\textbf{CLIP}} \\
    \cmidrule(r){1-3}\cmidrule(l){4-6}
    Original & 34.60 & 26.96 & 13.04 & 13.38 & 13.92 \\
    LIT      & 22.46 & 21.23 & 8.53 & 7.57 & 8.95 \\
    \cmidrule(r){1-3}\cmidrule(l){4-6}
    Reduction &
    \textcolor{green!60!black}{$\downarrow$\textbf{35.09\%}} &
    \textcolor{green!60!black}{$\downarrow$\textbf{21.25\%}} &
    \textcolor{green!60!black}{$\downarrow$\textbf{34.55\%}} &
    \textcolor{green!60!black}{$\downarrow$\textbf{43.40\%}} &
    \textcolor{green!60!black}{$\downarrow$\textbf{35.70\%}} \\
    \bottomrule
\end{tabular} } 
\vspace{-2mm} 
\label{tab:seg_fg_dataset_model} 
\end{table}
\paragraph{Adapt to other models and tasks}

A key strength of our method is its adaptation: it functions as a plug-and-play module that integrates into user-interactive systems without modifying the architecture or retraining the backbone. 

We first validate the adaptation on different VOS models. We apply our method to two recent SAM2-based models: DAM4SAM~\cite{dam4sam} and SAMURAI~\cite{samurai}, and evaluate them on the VOST dataset. As shown in Table~\ref{tab:seg_fg_dataset_model} (a), our approach consistently reduces the number of user corrections.

We further evaluate our method beyond video segmentation to fine-grained image classification. We use CLIP ViT-B/32 in a zero-shot configuration, converting classification into an online streaming annotation task. Images are grouped by the model’s initial predicted label and processed sequentially. We adopt top-3 accuracy to decide when user correction is required: if the ground-truth label does not appear in the top-3 predictions, a misclassification is flagged and the user provides the correct label. Each correction supervises an update to a lightweight LoRA module, enabling the model to quickly adapt to the observed error. As new images in the group stream in and error occurs, we first query the LoRA module; if the correct label appears within its top-3 predictions, we accept the prediction. Otherwise, the user corrects the label and the LoRA is updated again. This simulates a realistic annotation scenario where visually and semantically similar mistakes recur and the model continuously improves through user feedback.

We conduct experiments on CUB-200-2011~\cite{cub200}, Stanford Cars~\cite{stanford-car}, and SUN397~\cite{SUN397}, measuring the average number of user corrections required per class for classes that require at least five corrections. As shown in Table~\ref{tab:seg_fg_dataset_model} (b), our method consistently reduces the annotation burden by 35\% - 43\%. The results highlight that our LIT-LoRA not only transfers across model architectures, but also extends to tasks involving image–text alignment.

\subsection{Ablation Studies}

\paragraph{Number of Epochs}

\begin{figure}[t]
\centering

\begin{subfigure}[t]{0.4\linewidth}
    \vspace{3pt}
    \centering
    \includegraphics[width=\linewidth]{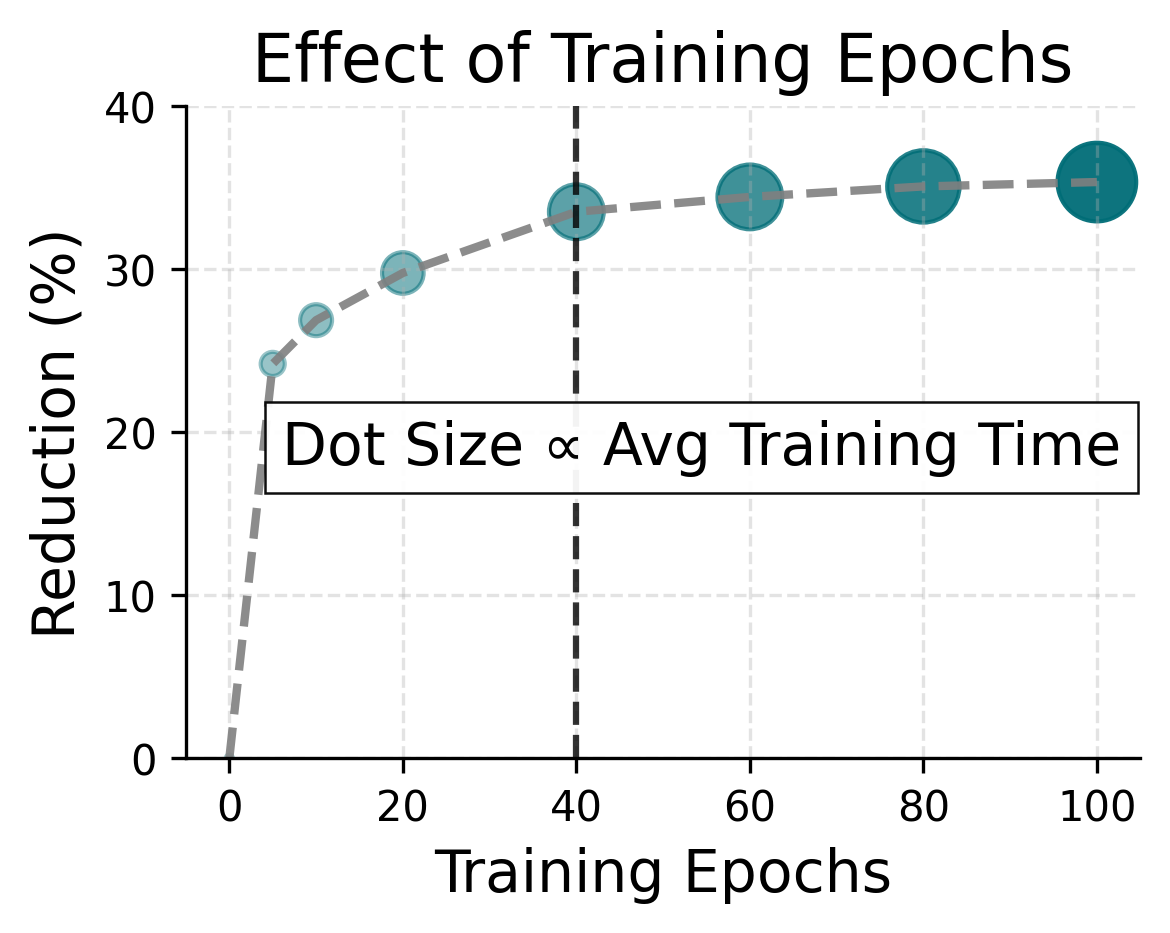}
    \vspace{-5mm}
    \caption{Effect of training epochs}
    \label{fig:epoch}
\end{subfigure}
\hspace{0.00\linewidth}
\begin{subfigure}[t]{0.58\linewidth}
    \vspace{10pt} 
    \centering
    \setlength{\tabcolsep}{0pt}
    \renewcommand{\arraystretch}{1.05}
    \resizebox{\linewidth}{!}{
    \begin{tabular}{lcc}
    \toprule
    \textbf{Method} & \textbf{\# Clicks} $\downarrow$ & \textbf{\#Params} \\
    \midrule
    Original (no adaptation) & 27.43 & --- \\
    Replace original & 32.47 & 35K \\
    LIT-LoRA (No CL) & 21.43 & 35K \\
    LIT-FT & 17.46 & 3.3M ($\sim$100$\times$) \\
    LIT-LoRA (3 LoRAs) & 17.87 & 105K (3$\times$) \\
    \textbf{LIT-LoRA (ours)} & \textbf{18.24} & \textbf{35K} \\
    \bottomrule
    \end{tabular}
    }
    \caption{Comparison of adaptation strategies}
    \label{tab:adaptation}
\end{subfigure}

\caption{\textbf{Ablation studies on VOST dataset}}
\label{fig:adaptation_epoch_table}
\end{figure}

We vary the number of training epochs used for LoRA training and report the reduction in user corrections along with the corresponding training time (Figure~\ref{fig:adaptation_epoch_table} (a)) on VOST. Increasing epochs improves correction reduction, but the gain quickly saturates: performance rises substantially from 5 to 40 epochs, while further increasing to 60–100 epochs yields only marginal improvement ($<$ 2 pp). In contrast, training latency grows roughly proportionally to the number of epochs. Considering this trade-off, we use 40 epochs in our experiments, which achieves near-optimal correction reduction while keeping the per-correction training overhead low.

\vspace{-4mm}

\paragraph{Different Design Choices}
We compare other adaptation strategies:
(1) \textit{Replace Original}: directly fine-tune the mask decoder at each correction, instead of maintaining a separate LIT-LoRA module;
(2) \textit{LoRA (No CL)}: fine-tune the LoRA only at the first correction, without continual learning for later corrections;
(3) \textit{LIT-FT}: fine-tune the entire mask decoder within the LIT framework instead of using LoRA module;
(4) \textit{LIT-LoRA (3-LoRAs)}: maintain up to three LoRA modules, where a new LoRA is initialized whenever the current one fails. For each new correction, the user selects the best prediction among the three, and only the selected LoRA is updated.

Figure~\ref{fig:adaptation_epoch_table}(b) summarizes correction efficiency and parameter cost. Simply finetuning the original decoder leads to worse performance, suggesting that naïve fine-tuning can overfit to errors and disrupt stable representations. Fine-tuning a single LoRA only once is also insufficient, confirming the importance of continual learning during interaction. While full fine-tuning and the 3-LoRA setup reduce clicks, they are substantially heavier, and the 3-LoRA approach additionally introduces cognitive overhead by requiring users to choose among multiple predictions.
Our design of LIT-LoRA achieves 18.24 clicks with only 35K parameters, providing the best balance between efficiency and usability.
\section{Conclusion}
\label{sec:conclusion}

We introduce Live Interactive Training (LIT), a framework for live model adaptation in user-interactive visual systems. Unlike models like SAM2, which receive feedback but cannot learn from it, LIT enables online updates during inference through lightweight and modular components. 
Our implementation, LIT-LoRA, instantiated in video segmentation on top of SAM2, effectively reduces user correction effort on challenging videos.
We further demonstrate the generality of LIT by applying it to multiple SAM2 variants and CLIP-based image classification tasks, achieving consistent improvements on correction efficiency across models and domains. We establish a step toward efficient, adaptive, and collaborative human–AI visual systems for complex real-world scenarios.

\section*{Acknowledgements}
This research is supported in part by the Cornell–LinkedIn Partnership and the National Science Foundation (IIS-2144117, IIS-2107161, and IIS-2505098). Yihong Sun is supported by an NSF Graduate Research Fellowship. Research supported by the NVIDIA Academic Grant Program using 4 $\times$ RTX PRO 6000 Blackwell GPUs.
{
    \small
    \bibliographystyle{ieeenat_fullname}
    \bibliography{main}

@String(CVPR= {IEEE Conf. Comput. Vis. Pattern Recog.})

@String(ICLR = {Int. Conf. Learn. Represent.})

@String(CVPR  = {CVPR})

@String(ICLR  = {ICLR})

@inproceedings{vscos,
  title={Video state-changing object segmentation},
  author={Yu, Jiangwei and Li, Xiang and Zhao, Xinran and Zhang, Hongming and Wang, Yu-Xiong},
  booktitle={Proceedings of the IEEE/CVF International Conference on Computer Vision},
  pages={20439--20448},
  year={2023}
}

@inproceedings{vost,
  title={Breaking the" object" in video object segmentation},
  author={Tokmakov, Pavel and Li, Jie and Gaidon, Adrien},
  booktitle={Proceedings of the IEEE/CVF Conference on Computer Vision and Pattern Recognition},
  pages={22836--22845},
  year={2023}
}

@inproceedings{sam,
  title={Segment anything},
  author={Kirillov, Alexander and Mintun, Eric and Ravi, Nikhila and Mao, Hanzi and Rolland, Chloe and Gustafson, Laura and Xiao, Tete and Whitehead, Spencer and Berg, Alexander C and Lo, Wan-Yen and others},
  booktitle={Proceedings of the IEEE/CVF international conference on computer vision},
  pages={4015--4026},
  year={2023}
}

@article{sam2,
  title={Sam 2: Segment anything in images and videos},
  author={Ravi, Nikhila and Gabeur, Valentin and Hu, Yuan-Ting and Hu, Ronghang and Ryali, Chaitanya and Ma, Tengyu and Khedr, Haitham and R{\"a}dle, Roman and Rolland, Chloe and Gustafson, Laura and others},
  journal={arXiv preprint arXiv:2408.00714},
  year={2024}
}

@article{samurai,
  title={Samurai: Adapting segment anything model for zero-shot visual tracking with motion-aware memory},
  author={Yang, Cheng-Yen and Huang, Hsiang-Wei and Chai, Wenhao and Jiang, Zhongyu and Hwang, Jenq-Neng},
  journal={arXiv preprint arXiv:2411.11922},
  year={2024}
}

@article{dam4sam,
  title={A Distractor-Aware Memory for Visual Object Tracking with SAM2},
  author={Videnovic, Jovana and Lukezic, Alan and Kristan, Matej},
  journal={arXiv preprint arXiv:2411.17576},
  year={2024}
}

@article{camsam2,
  title={CamSAM2: Segment Anything Accurately in Camouflaged Videos},
  author={Zhou, Yuli and Sun, Guolei and Li, Yawei and Fu, Yuqian and Benini, Luca and Konukoglu, Ender},
  journal={arXiv preprint arXiv:2503.19730},
  year={2025}
}

@inproceedings{sam2long,
  title={Sam2long: Enhancing sam 2 for long video segmentation with a training-free memory tree},
  author={Ding, Shuangrui and Qian, Rui and Dong, Xiaoyi and Zhang, Pan and Zang, Yuhang and Cao, Yuhang and Guo, Yuwei and Lin, Dahua and Wang, Jiaqi},
  booktitle={Proceedings of the IEEE/CVF International Conference on Computer Vision},
  pages={13614--13624},
  year={2025}
}

@InProceedings{rmem,
    author    = {Zhou, Junbao and Pang, Ziqi and Wang, Yu-Xiong},
    title     = {RMem: Restricted Memory Banks Improve Video Object Segmentation},
    booktitle = {Proceedings of the IEEE/CVF Conference on Computer Vision and Pattern Recognition (CVPR)},
    month     = {June},
    year      = {2024},
    pages     = {18602-18611}
}

@inproceedings{cutie,
  title={Putting the object back into video object segmentation},
  author={Cheng, Ho Kei and Oh, Seoung Wug and Price, Brian and Lee, Joon-Young and Schwing, Alexander},
  booktitle={Proceedings of the IEEE/CVF Conference on Computer Vision and Pattern Recognition},
  pages={3151--3161},
  year={2024}
}

@inproceedings{xmem,
  title={Xmem: Long-term video object segmentation with an atkinson-shiffrin memory model},
  author={Cheng, Ho Kei and Schwing, Alexander G},
  booktitle={European Conference on Computer Vision},
  pages={640--658},
  year={2022},
  organization={Springer}
}

@inproceedings{mivos,
  title={Modular interactive video object segmentation: Interaction-to-mask, propagation and difference-aware fusion},
  author={Cheng, Ho Kei and Tai, Yu-Wing and Tang, Chi-Keung},
  booktitle={Proceedings of the IEEE/CVF Conference on Computer Vision and Pattern Recognition},
  pages={5559--5568},
  year={2021}
}

@inproceedings{eva-vos,
  title={Learning the what and how of annotation in video object segmentation},
  author={Delatolas, Thanos and Kalogeiton, Vicky and Papadopoulos, Dim P},
  booktitle={Proceedings of the IEEE/CVF Winter Conference on Applications of Computer Vision},
  pages={6951--6961},
  year={2024}
}

@inproceedings{lazy-xeme,
  title={Strike the Balance: On-the-Fly Uncertainty based User Interactions for Long-Term Video Object Segmentation},
  author={Vujasinovi{\'c}, St{\'e}phane and Becker, Stefan and Bullinger, Sebastian and Scherer-Negenborn, Norbert and Arens, Michael and Stiefelhagen, Rainer},
  booktitle={Proceedings of the Asian Conference on Computer Vision},
  pages={2784--2802},
  year={2024}
}

@inproceedings{osvos,
  title={One-shot video object segmentation},
  author={Caelles, Sergi and Maninis, Kevis-Kokitsi and Pont-Tuset, Jordi and Leal-Taix{\'e}, Laura and Cremers, Daniel and Van Gool, Luc},
  booktitle={Proceedings of the IEEE conference on computer vision and pattern recognition},
  pages={221--230},
  year={2017}
}

@inproceedings{continuous,
  title={Continuous adaptation for interactive object segmentation by learning from corrections},
  author={Kontogianni, Theodora and Gygli, Michael and Uijlings, Jasper and Ferrari, Vittorio},
  booktitle={Computer Vision--ECCV 2020: 16th European Conference, Glasgow, UK, August 23--28, 2020, Proceedings, Part XVI 16},
  pages={579--596},
  year={2020},
  organization={Springer}
}

@article{hu2022lora,
  title={Lora: Low-rank adaptation of large language models.},
  author={Hu, Edward J and Shen, Yelong and Wallis, Phillip and Allen-Zhu, Zeyuan and Li, Yuanzhi and Wang, Shean and Wang, Lu and Chen, Weizhu and others},
  journal={ICLR},
  volume={1},
  number={2},
  pages={3},
  year={2022}
}

@article{bai2023qwen,
  title={Qwen technical report},
  author={Bai, Jinze and Bai, Shuai and Chu, Yunfei and Cui, Zeyu and Dang, Kai and Deng, Xiaodong and Fan, Yang and Ge, Wenbin and Han, Yu and Huang, Fei and others},
  journal={arXiv preprint arXiv:2309.16609},
  year={2023}
}

@article{cellseg1,
  title={CellSeg1: Robust Cell Segmentation with One Training Image},
  author={Zhou, Peilin and Du, Bo and Xu, Yongchao},
  journal={arXiv preprint arXiv:2412.01410},
  year={2024}
}

@article{sam2adapter,
  title={Sam2-adapter: Evaluating \& adapting segment anything 2 in downstream tasks: Camouflage, shadow, medical image segmentation, and more},
  author={Chen, Tianrun and Lu, Ankang and Zhu, Lanyun and Ding, Chaotao and Yu, Chunan and Ji, Deyi and Li, Zejian and Sun, Lingyun and Mao, Papa and Zang, Ying},
  journal={arXiv preprint arXiv:2408.04579},
  year={2024}
}

@article{convlora,
  title={Convolution meets lora: Parameter efficient finetuning for segment anything model},
  author={Zhong, Zihan and Tang, Zhiqiang and He, Tong and Fang, Haoyang and Yuan, Chun},
  journal={arXiv preprint arXiv:2401.17868},
  year={2024}
}

@inproceedings{cliplora,
  title={Low-rank few-shot adaptation of vision-language models},
  author={Zanella, Maxime and Ben Ayed, Ismail},
  booktitle={Proceedings of the IEEE/CVF Conference on Computer Vision and Pattern Recognition},
  pages={1593--1603},
  year={2024}
}

@misc{hong2024lvosbenchmarklargescalelongterm,
      title={LVOS: A Benchmark for Large-scale Long-term Video Object Segmentation}, 
      author={Lingyi Hong and Zhongying Liu and Wenchao Chen and Chenzhi Tan and Yuang Feng and Xinyu Zhou and Pinxue Guo and Jinglun Li and Zhaoyu Chen and Shuyong Gao and Wei Zhang and Wenqiang Zhang},
      year={2024},
      eprint={2404.19326},
      archivePrefix={arXiv},
      primaryClass={cs.CV},
      url={https://arxiv.org/abs/2404.19326}, 
}

@inproceedings{TTT,
  title={Test-time training with self-supervision for generalization under distribution shifts},
  author={Sun, Yu and Wang, Xiaolong and Liu, Zhuang and Miller, John and Efros, Alexei and Hardt, Moritz},
  booktitle={International conference on machine learning},
  pages={9229--9248},
  year={2020},
  organization={PMLR}
}

@article{TTTMAE,
  title={Test-time training with masked autoencoders},
  author={Gandelsman, Yossi and Sun, Yu and Chen, Xinlei and Efros, Alexei},
  journal={Advances in Neural Information Processing Systems},
  volume={35},
  pages={29374--29385},
  year={2022}
}

@article{TTTLLM,
  title={Test-time training on nearest neighbors for large language models},
  author={Hardt, Moritz and Sun, Yu},
  journal={arXiv preprint arXiv:2305.18466},
  year={2023}
}

@inproceedings{clip,
  title={Learning transferable visual models from natural language supervision},
  author={Radford, Alec and Kim, Jong Wook and Hallacy, Chris and Ramesh, Aditya and Goh, Gabriel and Agarwal, Sandhini and Sastry, Girish and Askell, Amanda and Mishkin, Pamela and Clark, Jack and others},
  booktitle={International conference on machine learning},
  pages={8748--8763},
  year={2021},
  organization={PmLR}
}

@article{videomae,
  title={Videomae: Masked autoencoders are data-efficient learners for self-supervised video pre-training},
  author={Tong, Zhan and Song, Yibing and Wang, Jue and Wang, Limin},
  journal={Advances in neural information processing systems},
  volume={35},
  pages={10078--10093},
  year={2022}
}

@article{online,
  title={Online deep learning: Learning deep neural networks on the fly},
  author={Sahoo, Doyen and Pham, Quang and Lu, Jing and Hoi, Steven CH},
  journal={arXiv preprint arXiv:1711.03705},
  year={2017}
}

@article{onlinemeta,
  title={Deep online learning via meta-learning: Continual adaptation for model-based RL},
  author={Nagabandi, Anusha and Finn, Chelsea and Levine, Sergey},
  journal={arXiv preprint arXiv:1812.07671},
  year={2018}
}

@article{ding2025mosev2,
  title={MOSEv2: A more challenging dataset for video object segmentation in complex scenes},
  author={Ding, Henghui and Ying, Kaining and Liu, Chang and He, Shuting and Jiang, Xudong and Jiang, Yu-Gang and Torr, Philip HS and Bai, Song},
  journal={arXiv preprint arXiv:2508.05630},
  year={2025}
}

@inproceedings{ding2023mose,
  title={MOSE: A new dataset for video object segmentation in complex scenes},
  author={Ding, Henghui and Liu, Chang and He, Shuting and Jiang, Xudong and Torr, Philip HS and Bai, Song},
  booktitle={Proceedings of the IEEE/CVF international conference on computer vision},
  pages={20224--20234},
  year={2023}
}

@article{cub200, 
title={The Caltech-UCSD Birds-200-2011 Dataset}, abstractNote={CUB-200-2011 is an extended version of CUB-200 [7], a challenging dataset of 200 bird species. The extended version roughly doubles the number of images per category and adds new part localization annotations. All images are annotated with bounding boxes, part locations, and at- tribute labels. Images and annotations were filtered by mul- tiple users of Mechanical Turk. We introduce benchmarks and baseline experiments for multi-class categorization and part localization.}, publisher={California Institute of Technology}, author={Wah, Catherine and Branson, Steve and Welinder, Peter and Perona, Pietro and Belongie, Serge}, year={2011}, month={Jul} }

@inproceedings{stanford-car,
  title={3d object representations for fine-grained categorization},
  author={Krause, Jonathan and Stark, Michael and Deng, Jia and Fei-Fei, Li},
  booktitle={Proceedings of the IEEE international conference on computer vision workshops},
  pages={554--561},
  year={2013}
}

@INPROCEEDINGS{SUN397,
  author={Xiao, Jianxiong and Hays, James and Ehinger, Krista A. and Oliva, Aude and Torralba, Antonio},
  booktitle={2010 IEEE Computer Society Conference on Computer Vision and Pattern Recognition}, 
  title={SUN database: Large-scale scene recognition from abbey to zoo}, 
  year={2010},
  volume={},
  number={},
  pages={3485-3492},
  doi={10.1109/CVPR.2010.5539970}}

@inproceedings{sam_hq,
    title={Segment Anything in High Quality},
    author={Ke, Lei and Ye, Mingqiao and Danelljan, Martin and Liu, Yifan and Tai, Yu-Wing and Tang, Chi-Keung and Yu, Fisher},
    booktitle={NeurIPS},
    year={2023}
}

@article{surgicalsam2,
  title={A fine-tuned foundational model SurgiSAM2 for surgical video anatomy segmentation and detection},
  author={Kamtam, Devanish N and Shrager, Joseph B and Malla, Satya Deepya and Wang, Xiaohan and Lin, Nicole and Cardona, Juan J and Yeung-Levy, Serena and Hu, Clarence},
  journal={Scientific Reports},
  volume={15},
  number={1},
  pages={35961},
  year={2025},
  publisher={Nature Publishing Group UK London}
}

@article{polysam2,
  title={Fine-Tuning SAM2 for Generalizable Polyp Segmentation with a Channel Attention-Enhanced Decoder},
  author={Liu, Yixiao},
  journal={Advanced Medical Research},
  volume={4},
  number={1},
  pages={1--9},
  year={2025}
}

@inproceedings{mocamask,
  title={Implicit motion handling for video camouflaged object detection},
  author={Cheng, Xuelian and Xiong, Huan and Fan, Deng-Ping and Zhong, Yiran and Harandi, Mehrtash and Drummond, Tom and Ge, Zongyuan},
  booktitle={Proceedings of the IEEE/CVF Conference on Computer Vision and Pattern Recognition},
  pages={13864--13873},
  year={2022}
}

@inproceedings{CAD,
  title={It’s moving! a probabilistic model for causal motion segmentation in moving camera videos},
  author={Bideau, Pia and Learned-Miller, Erik},
  booktitle={European Conference on Computer Vision},
  pages={433--449},
  year={2016},
  organization={Springer}
}

@article{rosa,
  title={Rosa: Accurate parameter-efficient fine-tuning via robust adaptation},
  author={Nikdan, Mahdi and Tabesh, Soroush and Crn{\v{c}}evi{\'c}, Elvir and Alistarh, Dan},
  journal={arXiv preprint arXiv:2401.04679},
  year={2024}
}

@article{peft,
  title={Parameter-efficient fine-tuning of large-scale pre-trained language models},
  author={Ding, Ning and Qin, Yujia and Yang, Guang and Wei, Fuchao and Yang, Zonghan and Su, Yusheng and Hu, Shengding and Chen, Yulin and Chan, Chi-Min and Chen, Weize and others},
  journal={Nature machine intelligence},
  volume={5},
  number={3},
  pages={220--235},
  year={2023},
  publisher={Nature Publishing Group UK London}
}

@article{peft-fourier,
  title={Parameter-efficient fine-tuning with discrete fourier transform},
  author={Gao, Ziqi and Wang, Qichao and Chen, Aochuan and Liu, Zijing and Wu, Bingzhe and Chen, Liang and Li, Jia},
  journal={arXiv preprint arXiv:2405.03003},
  year={2024}
}

@article{ni2024pace,
  title={Pace: Marrying generalization in parameter-efficient fine-tuning with consistency regularization},
  author={Ni, Yao and Zhang, Shan and Koniusz, Piotr},
  journal={Advances in Neural Information Processing Systems},
  volume={37},
  pages={61238--61266},
  year={2024}
}

@inproceedings{hu2023vl,
  title={Vl-pet: Vision-and-language parameter-efficient tuning via granularity control},
  author={Hu, Zi-Yuan and Li, Yanyang and Lyu, Michael R and Wang, Liwei},
  booktitle={Proceedings of the IEEE/CVF International Conference on Computer Vision},
  pages={3010--3020},
  year={2023}
}

@inproceedings{hu2023llm,
  title={Llm-adapters: An adapter family for parameter-efficient fine-tuning of large language models},
  author={Hu, Zhiqiang and Wang, Lei and Lan, Yihuai and Xu, Wanyu and Lim, Ee-Peng and Bing, Lidong and Xu, Xing and Poria, Soujanya and Lee, Roy},
  booktitle={Proceedings of the 2023 conference on empirical methods in natural language processing},
  pages={5254--5276},
  year={2023}
}

@misc{dinov3,
      title={DINOv3}, 
      author={Oriane Siméoni and Huy V. Vo and Maximilian Seitzer and Federico Baldassarre and Maxime Oquab and Cijo Jose and Vasil Khalidov and Marc Szafraniec and Seungeun Yi and Michaël Ramamonjisoa and Francisco Massa and Daniel Haziza and Luca Wehrstedt and Jianyuan Wang and Timothée Darcet and Théo Moutakanni and Leonel Sentana and Claire Roberts and Andrea Vedaldi and Jamie Tolan and John Brandt and Camille Couprie and Julien Mairal and Hervé Jégou and Patrick Labatut and Piotr Bojanowski},
      year={2025},
      eprint={2508.10104},
      archivePrefix={arXiv},
      primaryClass={cs.CV},
      url={https://arxiv.org/abs/2508.10104}, 
}

@article{dubey2024llama,
  title={The llama 3 herd of models},
  author={Dubey, Abhimanyu and Jauhri, Abhinav and Pandey, Abhinav and Kadian, Abhishek and Al-Dahle, Ahmad and Letman, Aiesha and Mathur, Akhil and Schelten, Alan and Yang, Amy and Fan, Angela and others},
  journal={arXiv e-prints},
  pages={arXiv--2407},
  year={2024}
}

@article{icedit,
  title={In-context edit: Enabling instructional image editing with in-context generation in large scale diffusion transformer},
  author={Zhang, Zechuan and Xie, Ji and Lu, Yu and Yang, Zongxin and Yang, Yi},
  journal={arXiv preprint arXiv:2504.20690},
  year={2025}
}

@inproceedings{onavos,
  title={Online adaptation of convolutional neural networks for the 2017 davis challenge on video object segmentation},
  author={Voigtlaender, Paul and Leibe, Bastian},
  booktitle={The 2017 DAVIS Challenge on Video Object Segmentation-CVPR Workshops},
  volume={5},
  number={6},
  year={2017}
}

@inproceedings{CTTA,
  title={Continual test-time domain adaptation},
  author={Wang, Qin and Fink, Olga and Van Gool, Luc and Dai, Dengxin},
  booktitle={Proceedings of the IEEE/CVF Conference on Computer Vision and Pattern Recognition},
  pages={7201--7211},
  year={2022}
}

@article{controlCTTA,
  title={Controllable continual test-time adaptation},
  author={Shi, Ziqi and Lyu, Fan and Liu, Ye and Shang, Fanhua and Hu, Fuyuan and Feng, Wei and Zhang, Zhang and Wang, Liang},
  journal={arXiv preprint arXiv:2405.14602},
  year={2024}
}

@article{shalev2012online,
  title={Online learning and online convex optimization},
  author={Shalev-Shwartz, Shai and others},
  journal={Foundations and Trends{\textregistered} in Machine Learning},
  volume={4},
  number={2},
  pages={107--194},
  year={2012},
  publisher={Now Publishers, Inc.}
}

@inproceedings{qian2024efficient,
  title={Efficient non-stationary online learning by wavelets with applications to online distribution shift adaptation},
  author={Qian, Yu-Yang and Zhao, Peng and Zhang, Yu-Jie and Sugiyama, Masashi and Zhou, Zhi-Hua},
  booktitle={Forty-first International Conference on Machine Learning},
  year={2024}
}

@inproceedings{atanyan2024continuous,
  title={Continuous adaptation for interactive segmentation using teacher-student architecture},
  author={Atanyan, Barsegh and Khachatryan, Levon and Navasardyan, Shant and Wei, Yunchao and Shi, Humphrey},
  booktitle={Proceedings of the IEEE/CVF Winter Conference on Applications of Computer Vision},
  pages={789--799},
  year={2024}
}

@inproceedings{liu2025magicquill,
  title={Magicquill: An intelligent interactive image editing system},
  author={Liu, Zichen and Yu, Yue and Ouyang, Hao and Wang, Qiuyu and Cheng, Ka Leong and Wang, Wen and Liu, Zhiheng and Chen, Qifeng and Shen, Yujun},
  booktitle={Proceedings of the Computer Vision and Pattern Recognition Conference},
  pages={13072--13082},
  year={2025}
}

@inproceedings{wang2024promptcharm,
  title={Promptcharm: Text-to-image generation through multi-modal prompting and refinement},
  author={Wang, Zhijie and Huang, Yuheng and Song, Da and Ma, Lei and Zhang, Tianyi},
  booktitle={Proceedings of the 2024 CHI Conference on Human Factors in Computing Systems},
  pages={1--21},
  year={2024}
}

@inproceedings{peng2024designprompt,
  title={Designprompt: Using multimodal interaction for design exploration with generative ai},
  author={Peng, Xiaohan and Koch, Janin and Mackay, Wendy E},
  booktitle={Proceedings of the 2024 ACM Designing Interactive Systems Conference},
  pages={804--818},
  year={2024}
}

@inproceedings{wei2023online,
  title={Online prototype learning for online continual learning},
  author={Wei, Yujie and Ye, Jiaxin and Huang, Zhizhong and Zhang, Junping and Shan, Hongming},
  booktitle={Proceedings of the IEEE/CVF international conference on computer vision},
  pages={18764--18774},
  year={2023}
}

@inproceedings{wei2025online,
  title={Online-lora: Task-free online continual learning via low rank adaptation},
  author={Wei, Xiwen and Li, Guihong and Marculescu, Radu},
  booktitle={2025 IEEE/CVF Winter Conference on Applications of Computer Vision (WACV)},
  pages={6634--6645},
  year={2025},
  organization={IEEE}
}

@inproceedings{he2025cl,
  title={CL-LoRA: Continual Low-Rank Adaptation for Rehearsal-Free Class-Incremental Learning},
  author={He, Jiangpeng and Duan, Zhihao and Zhu, Fengqing},
  booktitle={Proceedings of the Computer Vision and Pattern Recognition Conference},
  pages={30534--30544},
  year={2025}
}

@article{zhao2024videoprism,
  title={Videoprism: A foundational visual encoder for video understanding},
  author={Zhao, Long and Gundavarapu, Nitesh B and Yuan, Liangzhe and Zhou, Hao and Yan, Shen and Sun, Jennifer J and Friedman, Luke and Qian, Rui and Weyand, Tobias and Zhao, Yue and others},
  journal={arXiv preprint arXiv:2402.13217},
  year={2024}
}

@inproceedings{wang2022adamix,
  title={Adamix: Mixture-of-adaptations for parameter-efficient model tuning},
  author={Wang, Yaqing and Agarwal, Sahaj and Mukherjee, Subhabrata and Liu, Xiaodong and Gao, Jing and Hassan, Ahmed and Gao, Jianfeng},
  booktitle={Proceedings of the 2022 Conference on Empirical Methods in Natural Language Processing},
  pages={5744--5760},
  year={2022}
}

@inproceedings{sung2022vl,
  title={Vl-adapter: Parameter-efficient transfer learning for vision-and-language tasks},
  author={Sung, Yi-Lin and Cho, Jaemin and Bansal, Mohit},
  booktitle={Proceedings of the IEEE/CVF conference on computer vision and pattern recognition},
  pages={5227--5237},
  year={2022}
}

@article{li2024adapter,
  title={Adapter-x: A novel general parameter-efficient fine-tuning framework for vision},
  author={Li, Minglei and Ye, Peng and Huang, Yongqi and Zhang, Lin and Chen, Tao and He, Tong and Fan, Jiayuan and Ouyang, Wanli},
  journal={arXiv preprint arXiv:2406.03051},
  year={2024}
}

@article{sun2025tracking,
  title={Tracking and Understanding Object Transformations},
  author={Sun, Yihong and Yang, Xinyu and Sun, Jennifer J and Hariharan, Bharath},
  journal={Advances in Neural Information Processing Systems},
  year={2025}
}
}

\clearpage
\setcounter{page}{1}
\maketitlesupplementary
\appendix
\section{Implementation Details}
\label{sec:supp_detail}

\paragraph{Datasets}
We evaluate our method on four challenging VOS benchmarks: 

\textit{VOST}~\cite{vost} evaluates robustness under object breaking and large transformations, featuring frequent splits, deformations, and small fragments that challenge mask propagation. Results are reported on the validation split.

\textit{LVOSv2}~\cite{hong2024lvosbenchmarklargescalelongterm} is a long-term VOS benchmark emphasizing persistent tracking through occlusions and visually similar distractors, with many videos spanning hundreds of frames. We report results on the validation set.

\textit{MOSEv2}~\cite{ding2025mosev2} extends \textit{MOSEv1}~\cite{ding2023mose} to more complex real-world scenes with frequent object disappearance, occlusion, crowding, and adverse conditions. Since MOSE lacks public validation masks and MOSEv1 was already used to train SAM2, we evaluate on a filtered subset of \textit{MOSEv2} training samples that do not overlap with MOSEv1.

\textit{SA-V}~\cite{sam2} is a large-scale dataset from SAM2, covering diverse real-world scenes and object types to test model generalization. We evaluate on its validation and test splits.

\paragraph{LIT-LoRA on VOS} 
Throughout the experiments on VOS, we use the SAM2.1-Large checkpoint released on 2024-09-30. In our setup, the SAM2 backbone $\theta$ is kept fixed, and a lightweight LoRA module is inserted into the mask decoder. The decoder is composed of stacked two-way transformer blocks, and we augment each attention layer by modifying the query, key, and value projections $(W_Q, W_K, W_V)$ with a low-rank residual update
\begin{equation*}
W = W_0 + \Delta W, 
\qquad 
\Delta W = B A,
\end{equation*}
where $A \in \mathbb{R}^{r \times d}$ and $B \in \mathbb{R}^{d \times r}$ are the trainable LoRA matrices. The IoU prediction head and object-existence head remain frozen during training.

For inference, SAM2 produces several mask candidates, and we follow the default configuration and select the mask associated with the first mask token as the final prediction. Online updates are driven by user-provided corrections. Each correction minimizes a composite loss
\begin{equation*}
\mathcal{L}
= \lambda_{\text{focal}} \, \mathcal{L}_{\text{focal}}
+ \lambda_{\text{dice}} \, \mathcal{L}_{\text{dice}},
\end{equation*}
with a weighting ratio of $\lambda_{\text{focal}} : \lambda_{\text{dice}} = 20 : 1$ following SAM2's default parameters. This design enables the decoder-side LoRA parameters $\Delta W$ to adapt online while the remainder of SAM2 stays frozen.

The LoRA for VOS is configured with a rank of 4, $\alpha=4$, dropout of 0.1, and a learning rate of $1\times10^{-4}$.

\paragraph{LIT-LoRA on image classification} 
In the experiments on image classification, we use CLIP ViT-B/32. The full CLIP backbone $\theta$ is frozen, and a lightweight LoRA module is inserted in the feature space. We train the image encoder LoRA parameters while keeping the text encoder frozen.

At inference time, CLIP assigns each image a class based on similarity to all text prompts, where we use the text prompt ``a photo of x''. Whenever the ground-truth label does not appear in the top-$k$ predictions (e.g. $k=3$), a user correction is triggered and used to update the LoRA module. Each correction optimizes a cross-entropy loss over all class prompts, supplemented by a margin-based separation term and $L_2$ regularization on $\Delta W$. 
\begin{equation*}
\mathcal{L}
= \mathcal{L}_{\text{CE}}
\;+\;
\lambda_{\text{margin}}\, \mathcal{L}_{\text{margin}}
\;+\;
\lambda_{2}\, \lVert \Delta W \rVert_2^2.
\end{equation*}

The margin loss is defined as
\begin{equation*}
\mathcal{L}_{\text{margin}}
= \max\bigl(0,\; m - (s_y - s_{\max})\bigr),
\end{equation*}
where $m$ specifies the required separation between the ground-truth logit $s_y$ and the highest incorrect logit $s_{\max}$. We set $m = 10.0$, $\lambda_{\text{margin}}=0.1$, $\lambda_2=1\times10^{-4}$.

The LoRA in image classification is configured with a rank of 8, $\alpha=16$, dropout of 0.1, and a learning rate of $1\times10^{-4}$. We use a higher LoRA rank for image classification because the adapter operates only on a small feature-space projection in CLIP. This provides additional expressive capacity for fine-grained corrections at negligible computational cost, while preserving fast online updates.

\begin{figure*}[ht]
    \centering
    \includegraphics[width=1\linewidth]{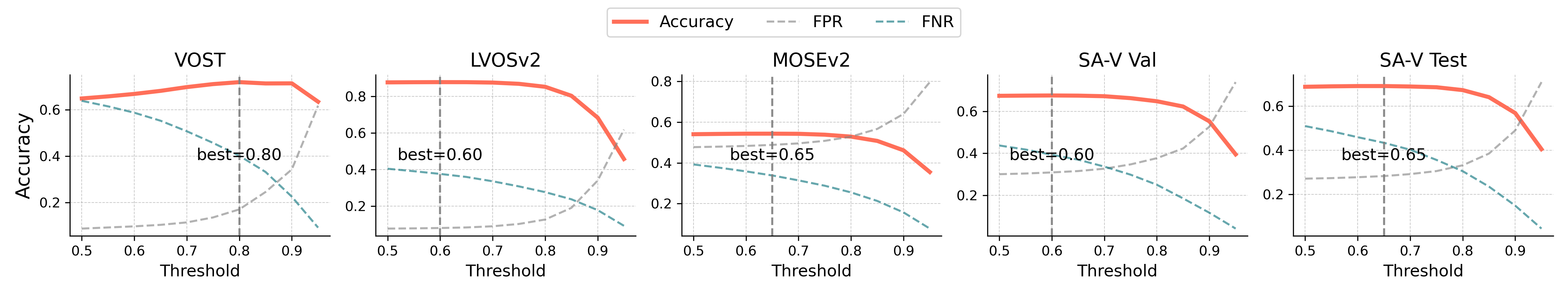}
    \caption{\textbf{Predicted IoU vs.\ Ground-Truth IoU (Accuracy, FPR, FNR).}
Accuracy, false positive rate (FPR), and false negative rate (FNR) across different predicted IoU thresholds, measuring how well the predicted IoU aligns with the ground-truth IoU for each dataset. Vertical dashed lines mark the threshold that achieves the highest alignment accuracy.}
    \label{fig:pred_iou}
\end{figure*}

\section{Additional Experiments and Analysis}

\paragraph{Robustness under different adapter insertion locations}

In our main experiment on VOS, we apply LoRA to the mask decoder, although other components also contribute to the final mask prediction. To assess the effectiveness of alternative lightweight modules, we additionally evaluate the use of an adapter on the memory bank. We freeze the mask decoder and insert a lightweight residual adapter module into the memory features. The adapter consists of two $1 \times 1$ convolutions with a ReLU in between, reducing the feature dimension by half and then restoring it, followed by a residual connection that adds the transformed features back to the input.

We compare the reduction in user corrections achieved by inserting the adapter at different locations within the SAM2 model on the VOST dataset. As shown in Table~\ref{tab:lora_location}, both insertion choices lead to consistent reductions in user corrections, with only minor differences in effectiveness. However, placing the adapter module in the memory features introduces more trainable parameters, resulting in higher computational overhead. Therefore, we adopt the mask-decoder insertion, which offers strong correction reduction while remaining lightweight.

\begin{table}[h]
\centering
\caption{\textbf{User-correction reduction with different adapter insertion locations.}}
\resizebox{\linewidth}{!}{
\begin{tabular}{lcc}
\toprule
 & \textbf{\# Clicks} $\downarrow$ & \textbf{\# Params} \\
\midrule
\textbf{Original} & 27.43 & -- \\
\textbf{$\text{LIT}_{\text{mask decoder}}$} & 18.24 & $\sim$35k \\
\textbf{$\text{LIT}_{\text{memory adapter}}$} & 18.84 & $\sim$66k ($\sim$2$\times$) \\
\bottomrule
\end{tabular}
}
\label{tab:lora_location}
\end{table}

\paragraph{Train on more corrections.} We compare training with more corrections under two batch size settings—joint updates ($bs=N$) and single-correction updates ($bs=1$)—against training with only one correction (Table~\ref{tab:corrections}). Joint updates ($bs=N$) degrade performance, as gradients from different corrections interfere under the limited capacity of LoRA adaptation. In contrast, sequential updates with $bs=1$ allow the adapter to incorporate each correction signal more effectively, improving performance at the cost of longer training time. Therefore, we adopt training with a single correction in our method.

\begin{table}[h]
\centering
\caption{\textbf{Train on more corrections.}}
\resizebox{\linewidth}{!}{
\begin{tabular}{lccc}
\toprule
 & \textbf{\# Clicks} $\downarrow$ & \textbf{Reduction} & \textbf{Time (s)} \\
\midrule
\textbf{Original} & 27.43 & -- & -- \\
\textbf{1/1 (ours)} & 18.24 & 33.51\% & 0.58 \\
\textbf{3/3} & 18.74 & 31.69\% & 0.61 \\
\textbf{5/5} & 19.03 & 30.62\% & 0.62 \\
\textbf{3/1} & 17.73 & 35.39\% & 1.61 \\
\textbf{5/1} & 17.64 & 35.71\% & 2.32 \\
\bottomrule
\end{tabular}
}
\label{tab:corrections}
\end{table}

\paragraph{Exploration of using predicted IoU from SAM2}

In our experiment, we assume a user is monitoring the video segmentation process and making decisions about when to apply corrections and whether the LIT-LoRA correction is acceptable. To move toward a more automated pipeline, we investigate whether model-internal signals can replace these human decisions for both error triggering and correction acceptance. 

SAM2’s mask decoder outputs include MLP heads that predict the IoU score and an occlusion score alongside the predicted masks, which could be a signal of the quality of the predicted mask. During training, SAM2 supervises the predicted IoU using ground truth IoU via an IoU output token. Motivated by this, we explore whether the predicted IoU can serve as a reliable signal for automatically detecting segmentation errors and determining whether a correction should be accepted.

To assess this, we evaluate how well the predicted IoU aligns with the ground truth IoU (set as 0.5) under different thresholds (Figure~\ref{fig:pred_iou}). Specifically, we define a prediction as correct when both the predicted and ground truth IoU values are simultaneously above or below a given threshold. This allows us to calculate an accuracy score for each threshold setting. We additionally report the false positive rate (FPR), where the predicted IoU overestimates mask quality, and the false negative rate (FNR), where it underestimates quality, to more precisely characterize this alignment behavior.

As shown in Figure~\ref{fig:pred_iou}, the predicted IoU score is unreliable in practice. Notably, the optimal threshold for predicted IoU to align with ground truth IoU could differ substantially from the predefined ground truth threshold of 0.5 used in our setup and varies across different datasets. For example, on VOST, the threshold that best aligns predicted IoU with the 0.5 ground-truth IoU is around 0.8, yet the accuracy peaks at only 0.74. Moreover, the accompanying FPR and FNR curves further highlight the instability of predicted IoU as a quality estimator: since predicted IoU scores are often underestimated, FNR is high at low thresholds, meaning many high-quality masks are incorrectly flagged as low quality; as the threshold increases, FPR rises quickly, causing the model to misclassify many low-quality masks as high quality and miss frames that actually require correction. This imbalance shows that predicted IoU does not provide a stable or trustworthy signal of true segmentation performance and should therefore be used cautiously. 

\begin{figure}[h]
    \centering
    \includegraphics[width=0.7\linewidth]{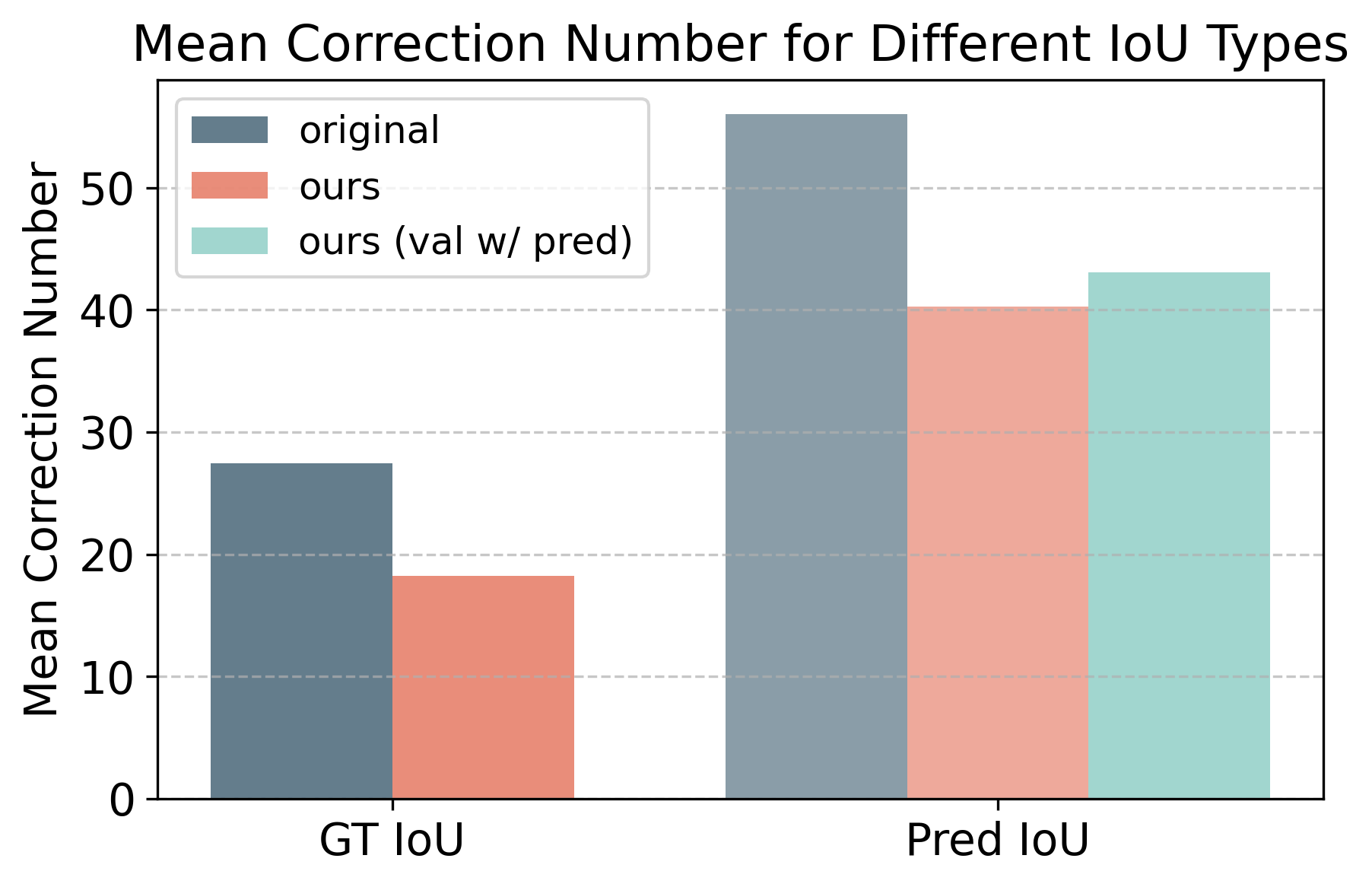}
    \caption{\textbf{Using predicted IoU for correction triggering and validation. }}
    \label{fig:pred_iou_vost}
\end{figure}

Nevertheless, we explore using the predicted IoU as a signal for both correction triggering and correction validation on VOST dataset. Based on our earlier analysis (Fig.~\ref{fig:pred_iou}), we adopt a threshold of 0.8 for predicted IoU in place of the 0.5 threshold defined for ground truth IoU ($\tau_{\mathrm{IoU}} = 0.5$). We evaluate two configurations:
(1) using predicted IoU only to trigger corrections, while still relying on the user (i.e., ground-truth IoU) to validate LIT-LoRA outputs (gray and red bars in Figure~\ref{fig:pred_iou_vost}); and
(2) using predicted IoU for both triggering and validation (green bar on the right in Figure~\ref{fig:pred_iou_vost}).

Our results (Figure~\ref{fig:pred_iou_vost}) show that although LIT-LoRA still reduces user effort under the automated settings, correction efficiency declines compared to full user supervision. Notably, the semi-automatic variant, where only triggering is automated using predicted IoU while validation remains user-based, achieves greater user effort reduction, but still lags behind the full user monitoring baseline. These findings highlight that while SAM2's predicted IoU can partially assist interaction, its misalignment with true segmentation quality limits its reliability. Developing a more accurate segmentation quality estimator remains a key avenue for enabling more effective automation.

\section{Limitations and Broader Impacts}
While our method provides an effective framework for efficient user corrections in interactive visual systems, it has some limitations.
First, our system requires user monitoring to detect errors, since SAM2 lacks a robust internal quality estimator. Incorporating automated validation mechanisms, such as learned IoU predictors or uncertainty estimation, could enable the system to automatically identify failure cases and further reduce the need for user intervention.
Second, our experiments primarily rely on synthetic user interactions. Although this protocol is standard in interactive segmentation research, real users may exhibit different behaviors and correction strategies. While we conducted a small-scale human evaluation to validate our method, a more systematic and large-scale user study would provide a more comprehensive assessment of real-world practice.
Third, our approach leverages LoRA-based live interactive training to achieve low-latency adaptation. This design assumes that the underlying base model already possesses strong generalization ability, enabling it to adapt quickly with lightweight updates. As a result, the effectiveness of our method may depend on large, well-pretrained models such as SAM2 or CLIP. In future work, we hope that advances in foundation models and more efficient adaptation mechanisms will further broaden the applicability of our framework to a wider range of visual systems.
\section{Additional Qualitative Results}

\begin{figure}[ht]
    \centering
    \includegraphics[width=\linewidth]{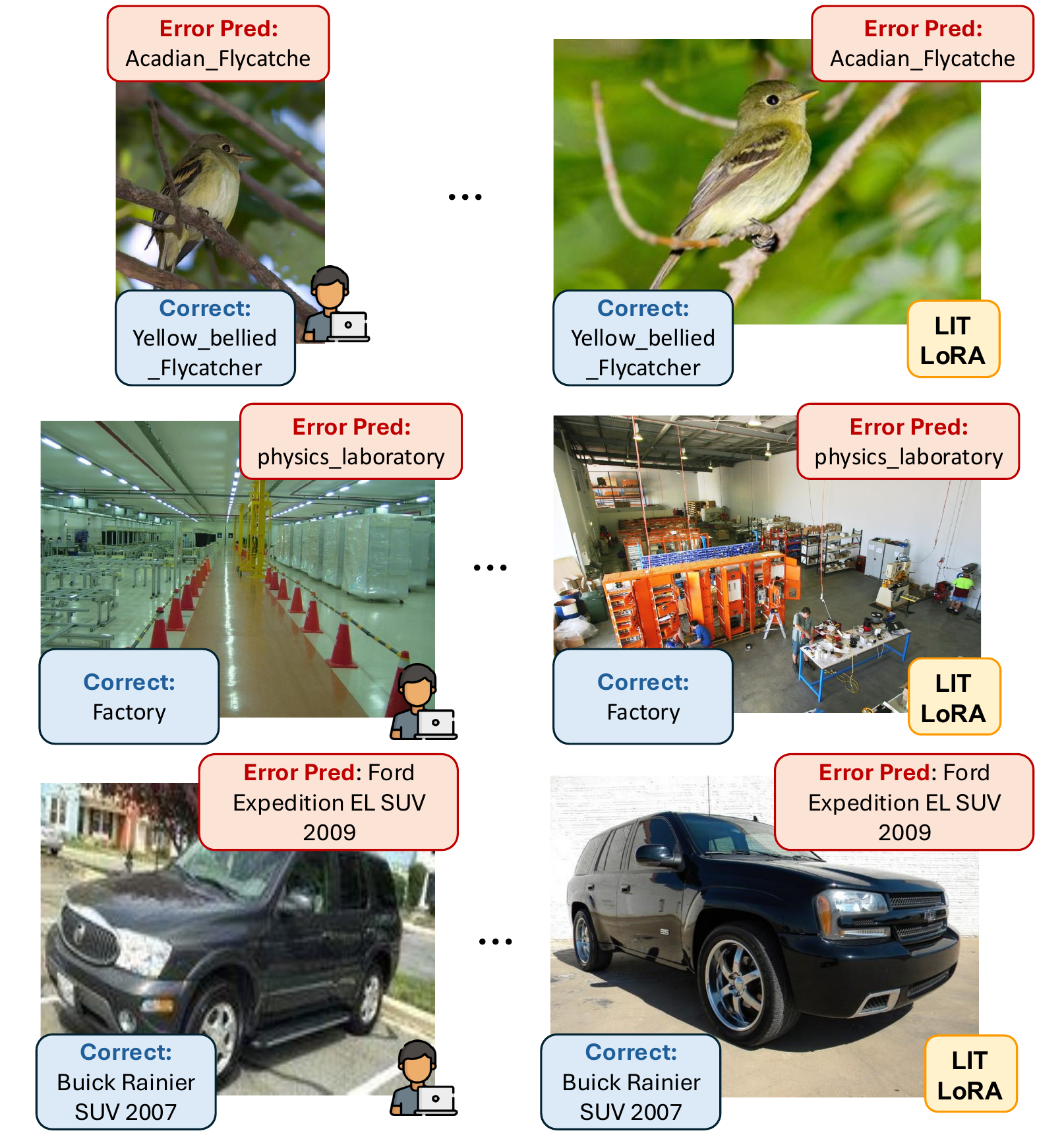}
    \caption{\textbf{Qualitative result of successful cases of LIT-LoRA on image classification.}}
    \label{fig:clip_qualitative}
\end{figure}

\paragraph{Successful cases} We present additional qualitative examples to highlight both the successful corrections and the broader applicability of LIT-LoRA. Figure~\ref{fig:clip_qualitative} shows successful cases on image classification: after learning from user-corrected labels, LIT-LoRA is able to correct similar mistakes that appear in subsequent images within the same group. This demonstrates that LIT-LoRA can effectively capture similar error patterns and correct them in future predictions, extending its usefulness beyond video object segmentation.

\begin{figure*}[h]
    \centering
    \includegraphics[width=1.0\linewidth]{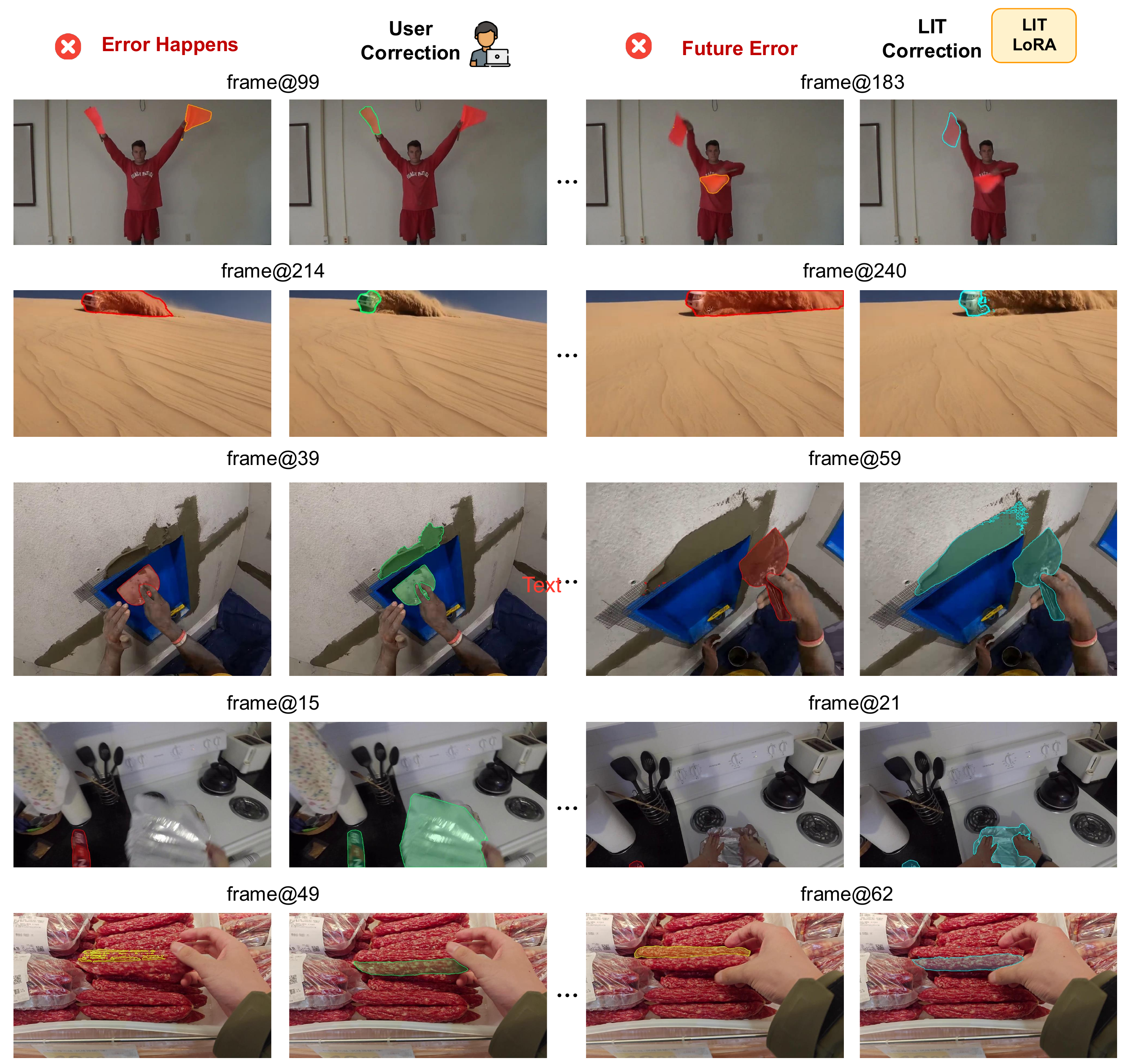}
    \caption{\textbf{Additional qualitative result of successful cases of LIT-LoRA on VOS.}}
    \label{fig:success}
\end{figure*}

We show additional successful LIT-LoRA adaptations on video object segmentation in Figure~\ref{fig:success}. These examples illustrate that LIT-LoRA can handle a broad range of challenging failure modes, including object switching, ambiguous boundaries, and complex object separations etc. Notably, it achieves these corrections from learning from a few prior user interactions, demonstrating strong responsiveness and the ability to refine segmentation quality under difficult visual and structural conditions.

\newpage
\begin{figure*}[htb]
    \centering
    \includegraphics[width=1.0\linewidth]{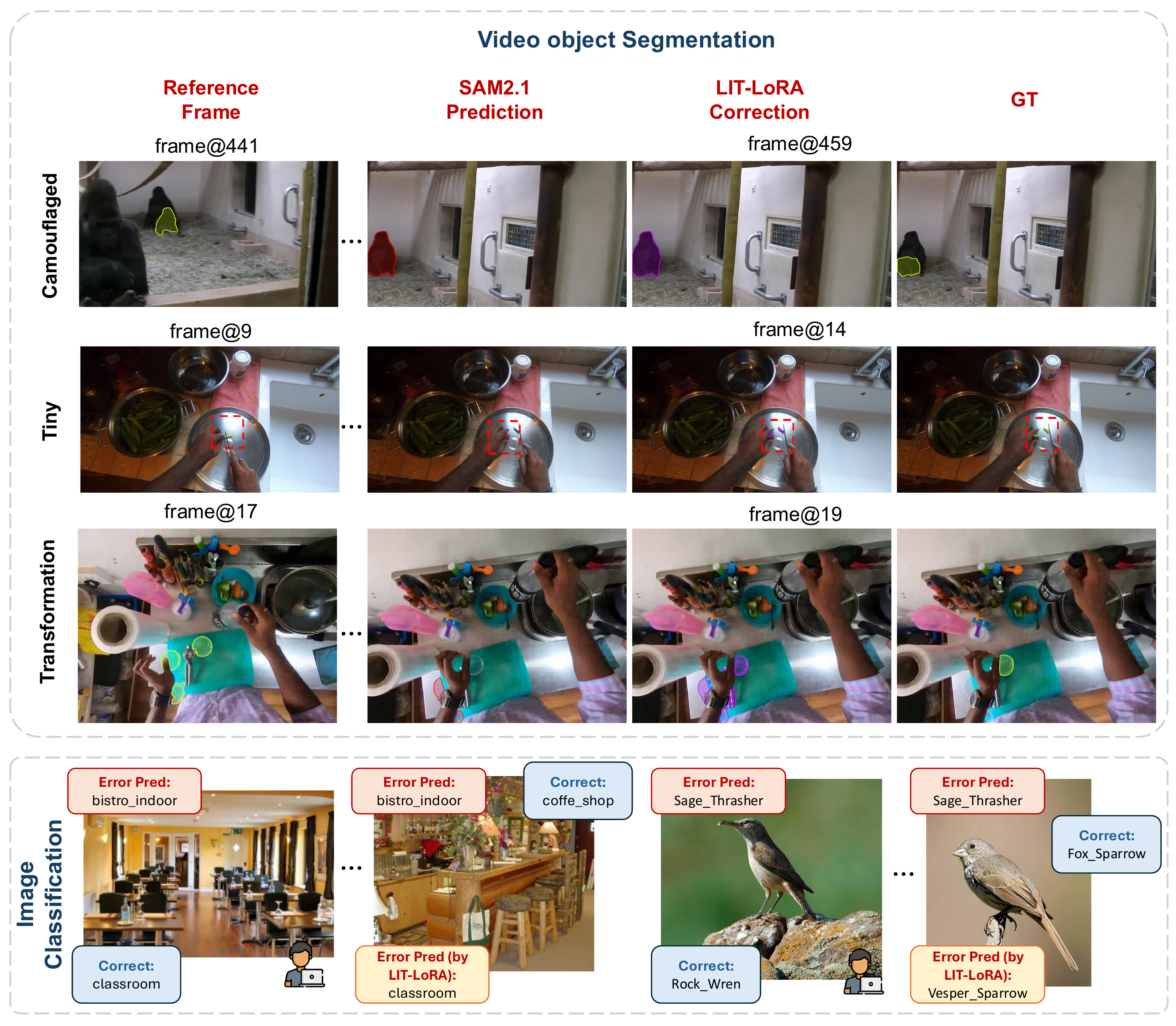}
    \caption{\textbf{Failure cases of LIT-LoRA.}} 
    \label{fig:failure}
\end{figure*}

\paragraph{Failure cases} We also present failure cases in Figure~\ref{fig:failure}, illustrating scenarios where LIT-LoRA is unable to correct the errors. In VOS tasks, the challenges include situations where the object is heavily camouflaged within a cluttered background (case 1), where the target object is extremely small (case 2), and where the object undergoes large or abrupt transformations that invalidate previously learned correction patterns (case 3). Such failures reveal limitations in the mask decoder’s generalization capacity and discrepancies between predicted and actual mask quality. Addressing these challenges may require richer training data or more robust adaptation mechanisms.

In the image classification task, LIT-LoRA can also face challenges. It may fail to correct mistakes when an image appears visually similar to past errors but actually belongs to a different class (case 1), or when the new error differs substantially from what it has previously learned (case 2). These difficulties are inherent to the online streaming setting, where new samples may diverge from earlier corrections. As more user-corrected examples accumulate and the model continues to adapt, LIT-LoRA can gradually become more stable and reliable.

Despite these limitations, LIT-LoRA remains highly effective in practice. It reduces user corrections by 18\%–34\% on VOS and 35\%–43\% on image classification, substantially reducing the user's annotation burden. Moreover, since users validate the corrections, occasional LIT-LoRA mistakes do not degrade final performance: incorrect updates can be rejected and corrected by the user, and these user-provided corrections further serve as supervision signals for continuous online learning. While we lay the foundation for more efficient and adaptive interactive visual systems, future work can explore strategies to further enhance the effectiveness and reliability of LIT-LoRA.

\end{document}